\documentclass{article}

\pdfoutput=1

\PassOptionsToPackage{numbers}{natbib}

\usepackage[preprint]{neurips_2024}

\usepackage[utf8]{inputenc} 
\usepackage[T1]{fontenc}    
\usepackage{hyperref}       
\usepackage{url}            
\usepackage{booktabs}       
\usepackage{amsfonts}       
\usepackage{nicefrac}       
\usepackage{microtype}      
\usepackage{xcolor}         
\usepackage{float}

\usepackage{bm} 
\usepackage{bbm} 
\usepackage{comment}
\usepackage{amsmath}
\usepackage{amssymb}
\usepackage{mathtools}
\usepackage{amsthm}
\usepackage{multirow}
\usepackage{algpseudocode}
\usepackage{algorithm}
\usepackage{algorithmicx}
\usepackage{algpseudocode}
\usepackage{setspace}
\usepackage{wrapfig}
\usepackage{subcaption}
\usepackage{graphicx}

\newcommand{\R}{\mathbb{R}}

\newcommand{\sood}{s}

\newcommand{\setid}{\mathcal{D}_{id}}
\newcommand{\setidfit}{\mathcal{D}_{id}^{fit}}
\newcommand{\setidtest}{\mathcal{D}_{id}^{test}}
\newcommand{\setideval}{\mathcal{D}_{id}^{eval}}
\newcommand{\setidval}{\mathcal{D}_{id}^{val}}
\newcommand{\setidcal}{\mathcal{D}_{id}^{cal}}
\newcommand{\setood}{\mathcal{D}_{ood}}
\newcommand{\setoodval}{\mathcal{D}_{ood}^{val}}
\newcommand{\setoodtest}{\mathcal{D}_{ood}^{test}}

\definecolor{lightergray}{rgb}{0.9, 0.9, 0.9}
\newcommand{\codevar}[1]{\colorbox{lightergray}{\texttt{#1}}}

\definecolor{beau-vert}{RGB}{0,102,51}

\everypar{\looseness=-1}

\def\eqref#1{equation~(\ref{#1})}

\def\1{\bm{1}}

\def\rvu{{\mathbf{i}}}

\def\rvs{{\mathbf{s}}}

\def\rvu{{\mathbf{u}}}

\def\rvx{{\mathbf{x}}}

\def\rvz{{\mathbf{z}}}

\def\vb{{\bm{b}}}

\def\vs{{\bm{s}}}

\def\vu{{\bm{u}}}

\def\vx{{\bm{x}}}
\def\vy{{\bm{y}}}

\def\mR{{\bm{R}}}

\DeclareMathAlphabet{\mathsfit}{\encodingdefault}{\sfdefault}{m}{sl}
\SetMathAlphabet{\mathsfit}{bold}{\encodingdefault}{\sfdefault}{bx}{n}

\def\sX{{\mathcal{X}}}

\def\prob#1{\mathcal{P}_{#1}}

\definecolor{pleasantgreen}{rgb}{0.2, 0.7, 0.2} 
\definecolor{pleasantred}{rgb}{1.0, 0.4, 0.4}   
\definecolor{pleasantgray}{gray}{0.5}           
\newcommand{\gain}[1]{%
  \ifdim #1pt > 0pt
    \small{\color{pleasantgreen}(+#1\%)}%
  \else
    \ifdim #1pt < 0pt
      \small{\color{pleasantred}(#1\%)}%
    \else
      \small{\color{pleasantgray}(+#1\%)}%
    \fi
  \fi
}

\RequirePackage[nolist,nohyperlinks]{acronym}
\begin{acronym}
\acro{ML}{{Machine Learning}}
\acro{DL}{{Deep Learning}}
\acro{OOD}{{Out-of-Distribution}}
\acro{ID}{{In-Distribution}}
\acro{CDF}{cumulative distribution function}
\end{acronym}

\newcommand{\multioodscore}{multi-dimensional OOD score}
\newcommand{\multidetector}{multi-dimensional detector}
\newcommand{\multiooddetector}{multi-dimensional OOD detector}

\title{Improving Out-of-Distribution Detection by Combining Existing Post-hoc Methods}

\author{%
  \textbf{Paul Novello}$^{1}$,
  \textbf{Yannick Prudent}$^{1}$,
  \textbf{Joseba Dalmau}$^{1}$,
  \textbf{Corentin Friedrich}$^{1}$,
  \textbf{Yann Pequignot}$^{2}$,\\
  $^{^1}$ DEEL, IRT Saint Exupery, \texttt{firstname.lastname@irt-saintexupery.com}\\
  $^{^2}$ Institut Intelligence des Données, Université de Laval, \texttt{yann.pequignot@iid.ulaval.ca}
}

\begin{document}

\maketitle

\begin{abstract}
Since the seminal paper of Hendrycks et al. \cite{hendrycks2016baseline}, Post-hoc deep Out-of-Distribution (OOD) detection has expanded rapidly. 
As a result, practitioners working on safety-critical applications and seeking to improve the robustness of a neural network now have a plethora of methods to choose from. 
However, no method outperforms every other on every dataset \cite{yang2022openood}, so the current best practice is to test all the methods on the datasets at hand. 
This paper shifts focus from developing new methods to effectively combining existing ones to enhance OOD detection.
We propose and compare four different strategies for integrating multiple detection scores into a unified OOD detector, based on techniques such as majority vote, empirical and copulas-based Cumulative Distribution Function modeling, and multivariate quantiles based on optimal transport. We extend common OOD evaluation metrics — like AUROC and FPR at fixed TPR rates — to these multi-dimensional OOD detectors, allowing us to evaluate them and compare them with individual methods on extensive benchmarks. Furthermore, we propose a series of guidelines to choose what OOD detectors to combine in more realistic settings, i.e. in the absence of known OOD data, relying on principles drawn from Outlier Exposure \cite{hendrycks2018deep}. The code is available at \url{https://github.com/paulnovello/multi-ood}.
\end{abstract}

\section{Introduction}\label{sec:intro}
Even though current \ac{ML} and \ac{DL} models are able to perform several complex
tasks that previously only human beings could (e.g. image classification, natural language processing),
we are still a step away from their widespread adoption, in particular in safety-critical applications.
One of the main reasons for the slow adoption of ML in safety-critical applications 
is the difficulty to certify a \ac{ML} component, 
mainly due to a poor control of the circumstances that may provoke such a ML component to fail.
In order to prevent failures of a \ac{ML} component, practitioners tend to avoid using the model
on data that differs greatly from the data used to train the model. Which bears the obvious question,
how do we distinguish between data that is similar to the training data, and data that is not?
\ac{OOD} detection tries to answer this question, and besides being a very active branch in machine learning research, \ac{OOD} detection has being recognized as an essential step in the certification of ML systems
by multiple certification authorities (see e.g. Sections 5.3 and 8.4 of \cite{Balduzzi2021neural}
or Section 5.1 of \cite{CoDANN2}).

Given the difficulty of training \ac{ML} and in particular \ac{DL} models,
post-hoc \ac{OOD} detection has emerged as the solution of choice for \ac{OOD} detection 
for most practitionners. Yet, the plethora of different existing \ac{OOD} detection methods to choose from
can be daunting to practitionners unfamiliar with the literature. 
Several recent solutions have made a great effort in order to foster accessibility 
to the field, e.g. surveys categorizing the different problems and associated methods \cite{yang2021generalized,ruff2021unifying} or libraries for benchmarking and comparing the existing methods \cite{yang2022openood, pytorchood, oodeel}. 
Nevertheless, the choice of different methods is still vast, 
and it is rather unclear from the literature what method one should choose for a given 
task or application, all the more since benchmarks tend to show that no method outperforms every other on every dataset.

In this paper, rather than trying to construct a new score and comparing it to existing ones,
we leverage the existing scores and ask ourselves how we can combine
the different scores in order to build a more powerful \ac{OOD} detector than
the ones built from individual scores. Our experiments show that there are 
multiple ways to combine individual scores, and that there is much to gain from doing so.
Our contributions can be summarized as follows:
\begin{itemize}
    \item We perform an empirical study of over 28 \ac{OOD} scores and show that several of them capture complementary information, while others capture redundant information. This suggests that there is room for improvement by combining scores, while also suggesting that only a subset of the individual scores considered will be useful in the combination scheme.
    \item We propose four different ways to combine \ac{OOD} scores: a simple majority vote, two methods based on the multi-dimensional \ac{CDF} (empirical \ac{CDF} and copulas), and multi-dimensional center-outward quantiles.
    \item We generalize common metrics for \ac{OOD} score evaluation, such as the AUROC and the FPR at fixed TPR to the multi-dimensional setting, thus allowing comparison between the new score combinations both with respect to the individual one-dimensional scores and with respect to each other.
    \item We perform an extensive empirical comparison and show that combining scores significantly improves upon state-of-the-art for \ac{OOD} detection.
    \item We propose two strategies to select the best score combination for a given problem or task:
    the first one in the presence of \ac{OOD} data, and the second one in the absence of \ac{OOD} data, by drawing inspiration from the outlier exposure technique \cite{hendrycks2018deep}.
\end{itemize}


\section{Related Works}
Most works that introduce score combinations for OOD detection take a single score function (e.g., Mahalanobis~\cite{lee_simple_2018} or DKNN~\cite{sun_out--distribution_2022}) and apply it across different layers of a neural network. 
This is the case of \cite{Darrin2024unsupervised}, where the aggregation is performed via simple operators like mean, max or min;
\cite{Himmi2024enhanced} where the different scores are combined using a weighted average, the weights being functions of both the score of the test example and the scores of the training examples; 
or yet \cite{Gisserotboukhlef2024trustworthy}, where they either combine the scores using a simple function like the maximum function, or if they have access to a reference dataset they fit a linear regression on the reference dataset in order to aggregate the scores.
More unusual ways of combining scores are given in \cite{Lin2021mood}, where a measure of the complexity of the test image is used in order to decide which layer to compute the score from; or \cite{Hogeweg2024cood}, where aggregation is performed by training a random forest classifier.
The work \cite{Colombo2022beyond} is slightly different, since aggregation of the feature vectors is performed before computing the score. This can be done thanks to the transformer architecture of the underlying network, which ensures that feature vectors have the same dimension for different layers.

Some works go further and combine scores obtained accross different layers of different neural networks. 
\cite{Jiang2023aggregating} build a score of the form: $s(x)=s_{cla}(x)+\lambda s_{rec}(x)$ by combining two scores for a same image: a score~$s_{cla}(x)$ obtained from the latent space of a classifier, and a score~$s_{rec}(x)$ obtained from the latent space of an autoencoder.
In a similar vein, \cite{Yang2021ensemble} build a score combining multiple Mahalanobis distances. The different distances are obtained from the feature spaces of the layers of a classifier network and a distance metric learning network. The scores are combined via a weighted sum, where the weights are fitted using a validation dataset that contains OOD samples. 
\cite{Bitterwolf2022breaking} approach OOD detection from a Bayesian perspective. 
Assuming there is an underlying classification task for ID data, they train both a classifier on the ID data and a binary discriminator on ID data and proxy OOD data (via e.g. outlier exposure). The classifier and the discriminator can either share the same latent representation or be trained independently.
The paper derives multiple OOD score functions that combine the output of the classifier and the discriminator.

Perhaps the closest work to our current approach is that of
\cite{Magesh2023principled}, where a series of hypothesis tests are written based on a different score each, then a text example is declared to be ID or OOD using a Benjamini-Hochberg type procedure (see \cite{Benjamini2001control}). They use conformal p-values in their hypothesis tests, thus obtaining a probabilistic guarantee. They combine both different score functions (Mahalanobis, Gram and Energy), as well as different scores obtained from the same score function (Mahalanobis or Gram) applied to different layers of the underlying neural network. 

We believe that the present work is substantially different to the existing literature: rather than using a single score function applied to different layers of the same neural network, we combine scores coming from multiple different score functions, many of which can be computed even in black-box settings where access to the feature vectors of the underlying neural network is unavailable. Moreover, our aggregation methods are novel compared to the current literature: instead of relying on simple operators like weighted averages or linear regressors, we leverage more elaborated mathematical tools like copulas or center-otward quantiles, which we believe are better suited to the problem of score aggregation. Finally, we perform an exhaustive study of score aggregation using 28 different score functions and testing the effect of the aggregation across 6 widely used benchmarks.

\vspace{-0.3cm}
\section{Setting}\label{sec:background-ood}
Given a probability distribution $\prob{id}$ on a space $\sX$
and a realization $\vx$ of a random variable $\rvx$ in $\sX$,
the task of \acf{OOD} detection consists in assessing 
if $\rvx\sim\prob{id}$ ($\rvx$ is considered \acf{ID}) 
or not ($\rvx$ is considered \ac{OOD}).
In most scenarios, the only information that one has about 
the distribution $\prob{id}$ is a set $\setid$ of i.i.d. realizations
drawn from the ID distribution.
The most common procedure for OOD detection is to construct a score function $\sood : \sX \rightarrow \R$ and a threshold $\tau\in\R$ such that: 
\begin{equation}
    \begin{cases}
        \ \rvx \;\;\text{is declared OOD} &\text{if}\ \sood(\vx) > \tau, \\
        \ \rvx \;\;\text{is declared ID} &\text{if}\ \sood(\vx) \leq \tau. \\
    \end{cases}
\end{equation}
We call $\sood$ an OOD score function. Usually the set $\smash{\setid}$ is split into
a set $\smash{\setidfit}$, which is used to construct the score function $\sood$,
and a set $\smash{\setideval}$ that is used for evaluation purposes.
Assume there are $p$ samples $\smash{\{\vx^{(1)}, \cdots, \vx^{(p)}\}}$ 
in the test dataset $\smash{\setideval}$.
To evaluate the quality of $\sood$, we consider $p$ additional OOD samples
$\smash{\{\bar{\vx}^{(1)},\cdots,\bar{\vx}^{(p)}\}}$ 
drawn from another distribution $\smash{\prob{ood} \neq \prob{id}}$ (typically, another dataset). 
We apply $\sood$ to obtain 
$\smash{\{\sood(\bar{\vx}^{(1)}), \cdots,\sood(\bar{\vx}^{(p)}),
\sood(\vx^{(1)}), \cdots, \sood(\vx^{(p)})\}}$.  
Then, we assess the discriminative power of $\sood$ by evaluating metrics depending on the threshold $\tau$. By considering ID samples as negative and OOD as positive, we can will mainly be using the Area Under the Receiver Operating Characteristic (AUROC): we compute the False Positive Rate (FPR) and the True Positive Rate (TPR) for 
      $\smash{\tau^i = \sood(\vx^{(i)})}$, 
      $i \in \{1, \cdots,p\}$, and compute the area under the curve with FPR as $x$-axis and TPR as $y$-axis.
  
In this work, we will be considering several OOD scores at the same time. For $d$ different score functions $s_1, \dots, s_d$, we denote 
by $\vs:\sX\rightarrow\R^d$ the \multioodscore{} function obtained by setting
$\vx \mapsto \vs(\vx)=(s_1(\vx),\dots, s_d(\vx))$.
In order to build a \multiooddetector{} from the score $\vs$,
we further split the set $\smash{\setideval}$ in three disjoint subsets:
$\smash{\setidcal}$, for calibrating the \multidetector{},
$\smash{\setidval}$, for comparing the different \multidetector{}s
to each other and chosing the best ones,
and $\smash{\setidtest}$, for final evaluation purposes.
  
\section{OOD Scores Combination}
\begin{figure}[!h]
  \centering
  \includegraphics[width=0.32\linewidth]{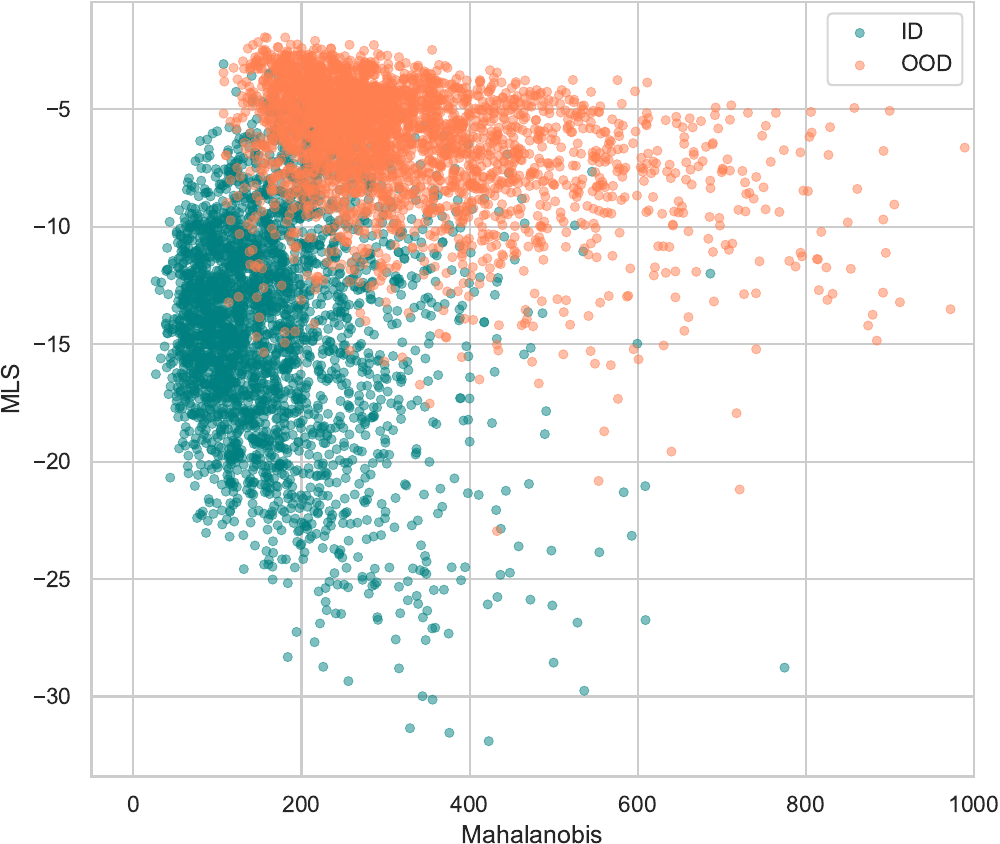}
  \includegraphics[width=0.32\linewidth]{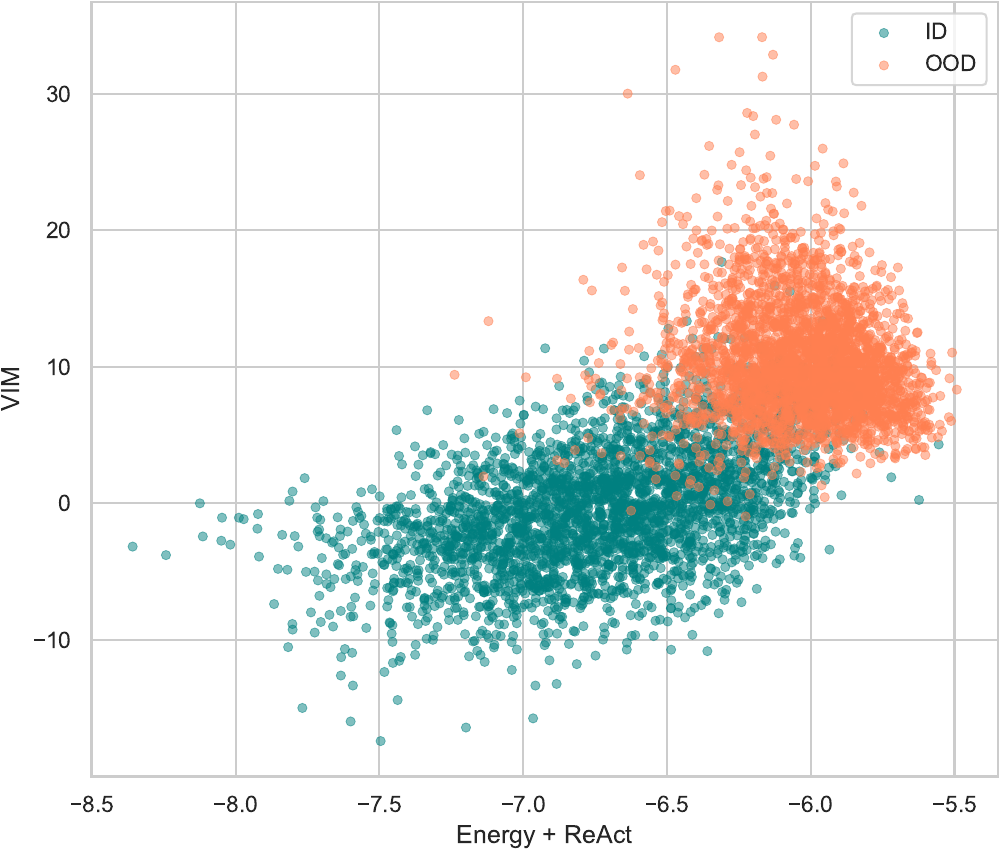}
  \includegraphics[width=0.32\linewidth]{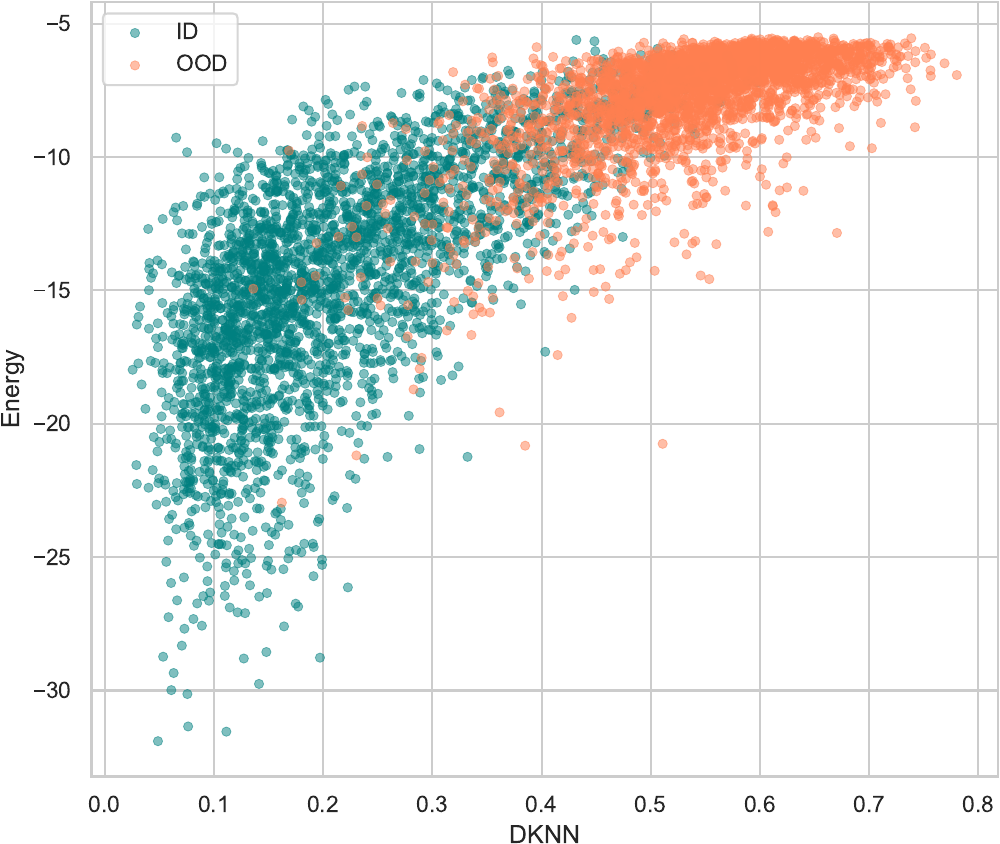}
    \caption{Plot of ID (Imagenet-200) and OOD (Textures) scores for three selected OOD score functions. The scores are not always correlated (we picked three that are not for illustration purposes), suggesting that a more sophisticated decision boundary could be built and improve OOD detection.   \label{fig:scatter}}
    \vspace{-0.5cm}
\end{figure}

The main motivation for our combining scores is the empirical observation that 
different \ac{OOD} score functions capture different information. This is obvious from Figure~\ref{fig:scatter},
where the two-dimensional point clouds are seen to better separate \ac{ID} from \ac{OOD}
data than the respective marginals. 

\subsection{Combination Methods}\label{sec:baselines}
We propose here four different methods for combining different score functions
into a single \multiooddetector{}. Assume we are given $d$ \ac{OOD} score functions
$s_1,\dots,s_d$, and the \multioodscore{} $\vs$ is constructed as in section~\ref{sec:background-ood}.
A \multiooddetector{} is defined by choosing a subset 
$A \subset \R^d$ and setting
\begin{equation}
    \begin{cases}
        \rvx \;\;\text{is declared OOD}\;\; &\text{if}\ \vs(\vx) \in A \\
        \rvx \;\;\text{is declared ID}\;\; &\text{if}\ \vs(\vx) \not\in A  \\
    \end{cases}
\end{equation}
We only consider \multiooddetector{}s that are \emph{monotonous} in the following sense:
$\smash{\1\{(s_1,\dots,s_d)\in A\}}$ is a monotonous function of $s_k$ for every $k$ when all other coordinates
are kept fixed. Note that this is a natural constraint, since each individual score $s_k$ is itself
an \ac{OOD} score.

\subsubsection{Majority Vote}
The majority vote combination approach is the most naive one: the individual binary decisions $s_i(\vx) \geq \tau_i$ (where $\tau_i$ is the threshold for the single OOD detector $i$) are aggregated through a simple majority voting scheme, where the final classification is determined by the most popular choice among the individual OOD detectors. 
In case where an equal number of detectors vote for ID and OOD (a possibility when the number of methods is even), we adopt a tie-breaking rule that favors OOD attribution. For example, if four OOD detectors assign two ID and two OOD predictions, the decision is OOD. Mathematically, it corresponds to choosing
\begin{equation}
    A = \bigg\{ \vs \in \R^d \;\; \bigg\rvert \;\; \sum_{i=1}^d \1\left(s_i\geq \tau_i\right) \geq \frac{d}{2} \bigg\}.
\end{equation}
For each OOD detector, a suitable threshold $\tau_i$ for binary classification must be determined. In practice, this threshold is typically computed by identifying the quantile that corresponds to a specific false positive rate on ID data, e.g. 5\% FPR. Like for single OOD detection evaluation, performance metrics - including AUROC, FPR95TPR, and TPR5FPR - can be calculated for the majority vote approach by systematically varying the common quantile used to determine individual thresholds.

\subsubsection{Empirical Cumulative Distribution Function}
The next two methods are based on the estimation of the multi-dimensional Cumulative Distribution Function (CDF) of the considered scores. 
Given an ID random variable $\smash{\rvx\sim\prob{id}}$, its multi-dimensional score $\vs(\rvx)$ is a random
vector in $\R^d$, we can define the CDF of $\vs(\rvx)$ as being the function
$\smash{F_{\vs(\rvx)}:\R^d\rightarrow\R}$ such that 
$\smash{F_{\vs(\rvx)}(\vy)=P(s_1(\rvx)\leq y_1,\dots, s_d(\rvx)\leq y_d)}$. 
The CDF is a natural way to combine scores because it can naturally take several scores as input and output a scalar.  
A first way to do so is to nonparametrically estimate $F_{\vs(\rvx)}$ using the empirical CDF function:
\begin{equation}
  \hat{F}_{\vs(\rvx)}(\vy) \,=\, \frac{1}{|\setidcal|}\,\sum_{\vx\in\setidcal} \1 \big(s_1(\vx) \leq y_1, \dots, s_d(\vx) \leq y_d \big).
\end{equation}
In order to build a \multiooddetector{} we pick a threshold $\tau$ in $(0,1)$ and define
\begin{equation}
    A \,=\, \big\{ \vy\in \R^d \;\; \big\rvert \;\; 
    \hat{F}_{\vs(\rvx)}(\vy) > \tau \big\}.
\end{equation}

\subsubsection{Parametric Cumulative Distribution Function with Copulas}
Another way of estimating $\smash{F_{\vs(\rvx)}}$ is by using a parametric estimator.
Let us consider the marginal CDFs of $\vs(\rvx)$, 
namely $\smash{F_{s_1(\rvx)}, ..., F_{s_d(\rvx)}}$. Each $\smash{F_{s_k(\rvx)}}$ can be estimated with 
$\smash{F^{\theta_k}:\R\mapsto [0,1]}$ 
chosen from a family of a known probability distribution, e.g., Gaussian, Beta, etc., of parameter $\theta_k$, the parameter $\theta_k$ is inferred from the ID scores $\smash{\setidcal}$. 
For each choice of the family of distributions, the parameter $\theta_k$ 
is fitted on $\smash{\big\lbrace s_k(\vx) \,:\, \vx\in\setidfit \big\rbrace}$ 
using Maximum Likelihood Estimation (MLE).

However, as shown by the correlations between scores in the Figure \ref{fig:scatter}, 
the scores might be dependent, so it is not enough to model the marginal CDFs $\smash{F^{\theta_k}}$;
the dependence between the random variables $s_k(\rvx)$ has to be taken into account. 
This can be achieved by considering copulas. 
Let consider a copula function $\smash{C^{\theta'}: [0,1]^d \rightarrow \R}$. It is possible to model the joint CDF of $\vs(\rvx)$ when they are not independent by $\smash{\hat{F}^{\theta}}$ defined as:
\begin{equation}
  \hat{F}^{\theta}(\vx) = C^{\theta'}\big(F^{\theta_1}(\vx),...,F^{\theta_d}(\vx) \big),
\end{equation}
where $\theta = (\theta_1,\dots,\theta_d,\theta')$.
The copula function can be chosen among a set of parametric copula families, one of the most common being archimedean copulas, and the parameter $\theta'$, can also be inferred from 
$\smash{\{\vs(\vx) \,:\, \vx \in\setidcal\}}$ using MLE.
As for the case of the empirical CDF, we build \multiooddetector{} by choosing
a threshold $\tau$ and defining
\begin{equation}
    A = \big\{ \vy\in \R^d \;\; \big\rvert \;\; 
    \hat{F}^\theta(\vy) > \tau \big\}.
\end{equation}

\subsubsection{Center-Outward Quantiles}
Center-outward quantiles, introduced in \cite{hallin2021distribution}, generalize univariate quantiles to multivariate settings, extending distribution and quantile functions to higher dimensions. This concept, rooted in optimal transportation theory, addresses the challenge of the lack of a canonical ordering in multidimensional spaces. In the context of combining OOD scores, we use this approach to estimate quantiles for the \multioodscore{}. A new sample is considered as ID for a fixed decision threshold $q$ if it can be matched to an ID quantile inferior to $q$; otherwise, the sample is considered as OOD. Here are the key steps of this approach:

\paragraph{Step 1: Generating Points on Nested Hyperspheres}
To represent the source distribution for the optimal transport problem, we generate $N$ points uniformly on the surfaces of $k$ nested hyperspheres in $\smash{\mathbb{R}^d}$,
intersected with the first quadrant $\smash{\R_+^d}$. 
Each point $\rvz^{(i)}$ is sampled from one of the hyperspheres with radius $r_t, t = 1, \dots, k$, such that:
\begin{equation}
\rvz^{(i)} = r_{t_i} \cdot \rvu^{(i)}, \quad i = 1, \ldots, N,
\end{equation}
where $\rvu^{(i)} \in \mathbb{R}^d$ are points uniformly distributed on 
$\smash{\mathbb{S}^{d-1}\cap\R_+^{d}}$ 
($\smash{\mathbb{S}^{d-1}}$ being the unit hypersphere in $\smash{\R^d}$), 
and $t_i$ 
are uniformly distributed over the set $\{1, \ldots, k\}$. The radius $r_t$ is scaled linearly such that $r_t = t/k$.

\paragraph{Step 2: Optimal Transport Plan}
Assume that the set $\smash{\setidcal}$ is of size $M$
and denote by $\smash{\vs^{(1)},\dots,\vs^{(M)}\in\R^d}$ their scores.
Given the set of reference points $\rvz^{(i)}$ on the hyperspheres 
we compute the optimal transport plan 
$\smash{\mathbf{P} \in \mathbb{R}_+^{N \times M}}$ 
that minimizes the transportation cost:
\begin{equation}
\mathbf{P} = \arg \min_{\mathbf{P} \in \Pi(\mathbf{a}, \mathbf{b})} \sum_{i=1}^{N} \sum_{j=1}^{M} \mathbf{P}_{ij} \cdot c\big(\rvz^{(i)}, \rvs^{(j)}\big),
\end{equation}
where $\smash{\Pi(\mathbf{a}, \mathbf{b})}$ 
is the set of all transportation plans that satisfy the marginal constraints $\mathbf{a}$ and $\mathbf{b}$ (uniform distributions over $\smash{\lbrace 1,\dots, N\rbrace}$ and 
$\smash{\lbrace 1,\dots,  M\rbrace}$, respectively), 
and 
$\smash{c\big(\rvz^{(i)}, \vs^{(j)}\big)}$ 
is the cost function, typically the squared Euclidean distance 
$\smash{c(\rvz, \vy) = \|\rvz - \vy\|_2^2}$.
Informally speaking, the meaning of $\mathbf{P}$ is that a mass $\smash{\mathbf{P}_{ij}}$ is transported from the point
$\rvz^{(i)}$ to the point $\vs^{(j)}$, given that each point $\rvz^{(i)}$ in the initial distribution has a mass of $1/N$
and each point $\vs^{(j)}$ has a mass of $1/M$.

\paragraph{Step 3: Estimating Quantiles}
The estimated quantile values $\mathbf{q}$ for the in-distribution scores are derived from the transport plan $\mathbf{P}$ and the radii of the hyperspheres. The quantile value for each score 
$\vs^{(j)}$ 
is given by:
\begin{equation}
q_j = \sum_{i=1}^{N} r_{t_i} \cdot \mathbf{P}_{ij},
\end{equation}
where $r_{t_i}$ corresponds to the radius of the hypersphere from which the point $\rvz^{(i)}$ was sampled.

\paragraph{Step 4: Constructing Quantile Contours}
To visualize and interpret the quantiles, we construct a convex hull for the points below a specific quantile threshold $Q$. The convex hull is defined as:
\begin{equation}
\text{Conv}\Big(\big\lbrace \vs^{(j)} \,\big|\, q_j \leq Q\big\rbrace \Big),
\end{equation}
where $Q$ is the desired quantile level.

\paragraph{Step 5: OOD classification}
\label{sec:co_ood_classif}
In order to build an \multiooddetector{} that satisfies the monotonicity condition
from section~\ref{sec:baselines}, 
we choose $A$ to be the complement of the smallest set
that contains the convex hull for which the monotonicity condition is satisfied.
The complement of $A$ can be built
by computing the convex hull of an extended set of points,
where new well-chosen points have been added to the set of ID scores.
New data points can then be classified by determining whether they lie inside the set $A$, which, thanks to the complement of $A$ being a convex hull, can be achieved using the Delaunay triangulation \cite{guibas1992randomized}. For a high-dimensional \multiooddetector{}, the computation of Delaunay triangulation might be prohibitive, so we also consider another variant based on K-Nearest-Neighbors, whose description is detailed in Appendix \ref{app:comb}. 




\subsection{OOD Metrics for \multiooddetector{}s}
\begin{figure*}[!t]
  \centering
  \includegraphics[width=0.23\textwidth]{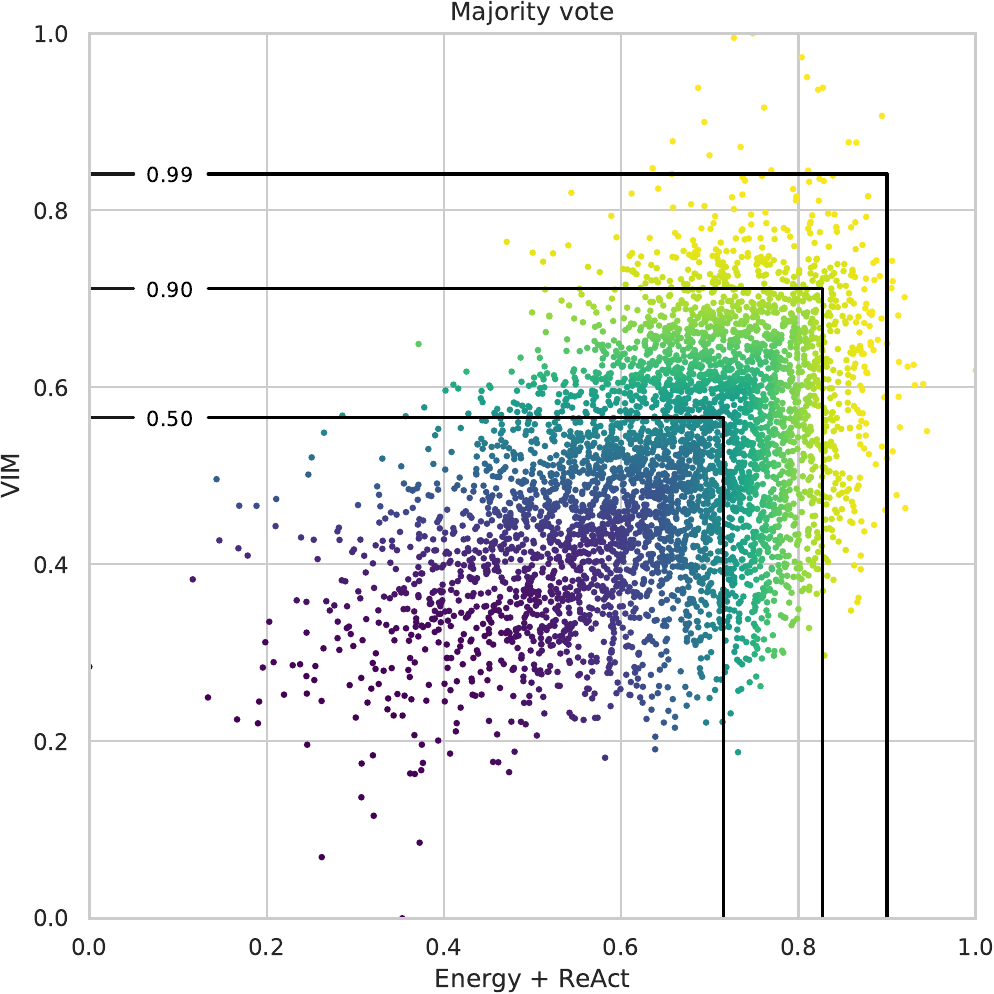}
  \includegraphics[width=0.23\textwidth]{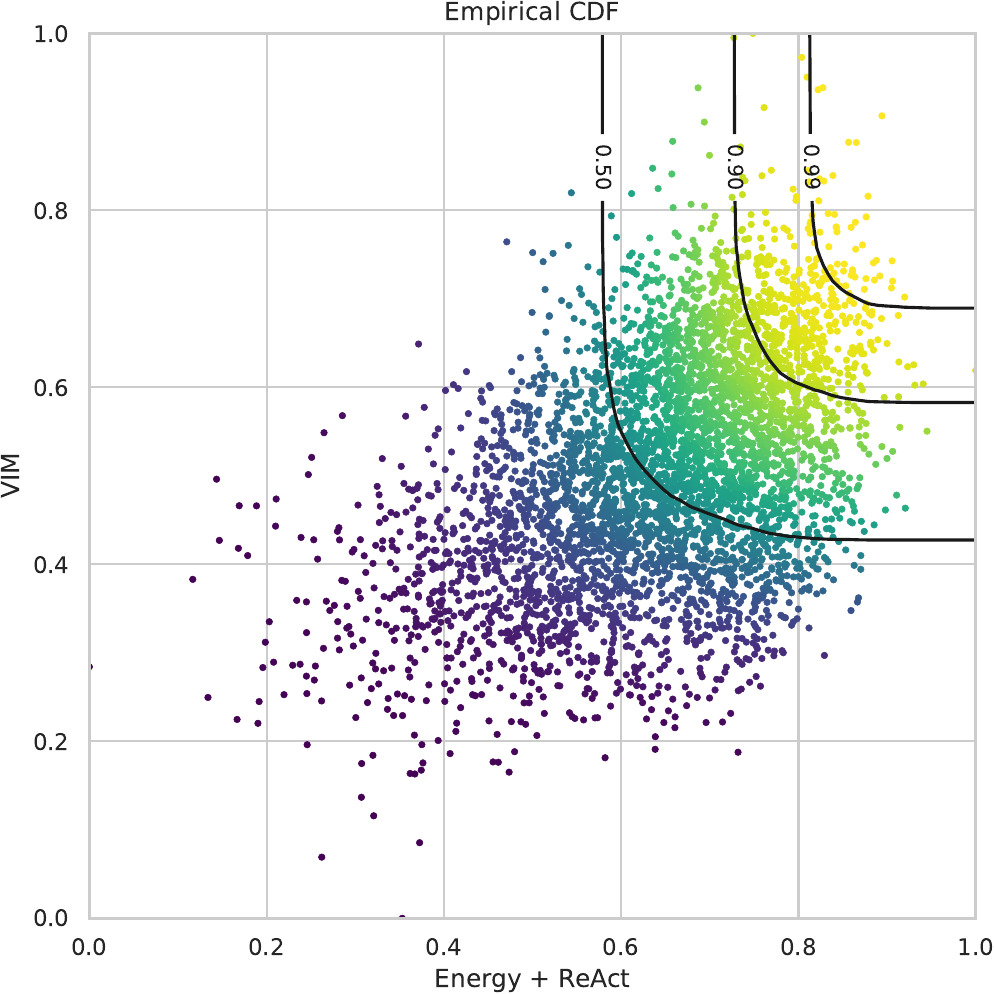}
  \includegraphics[width=0.23\textwidth]{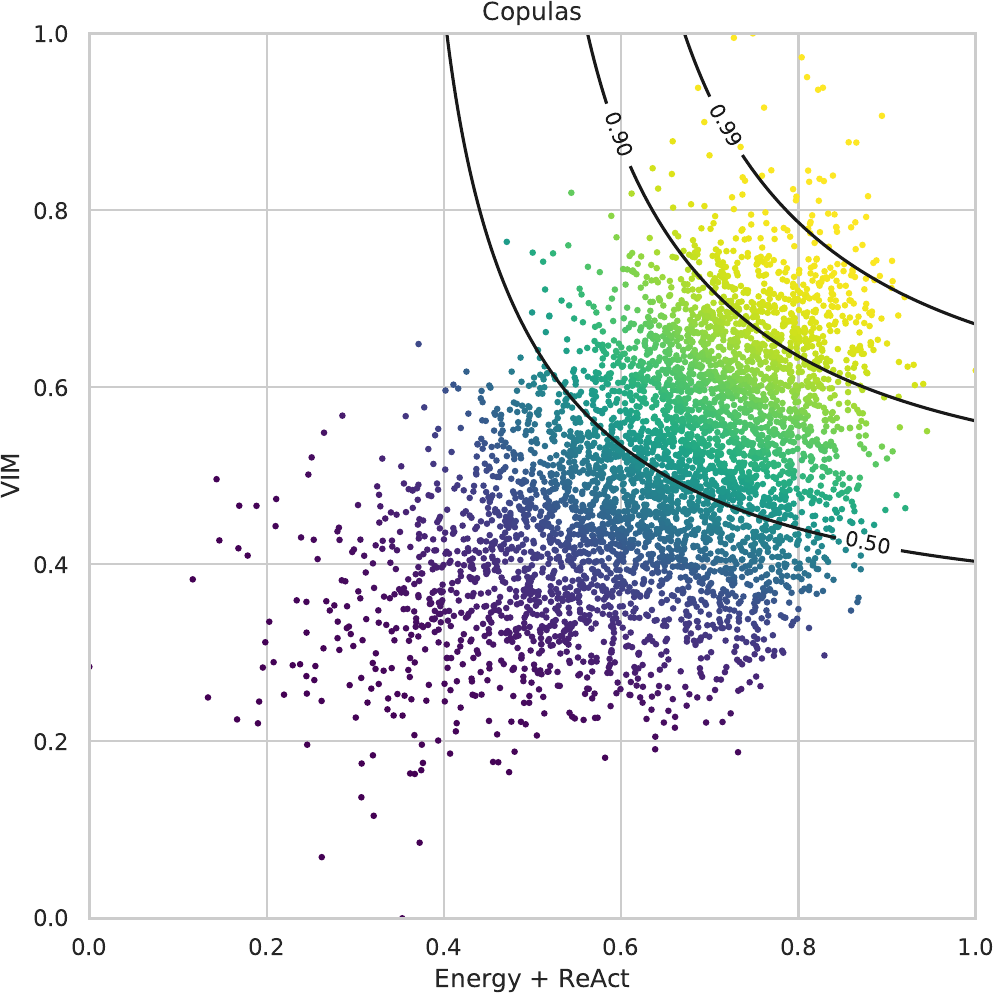}
  \includegraphics[width=0.28\textwidth]{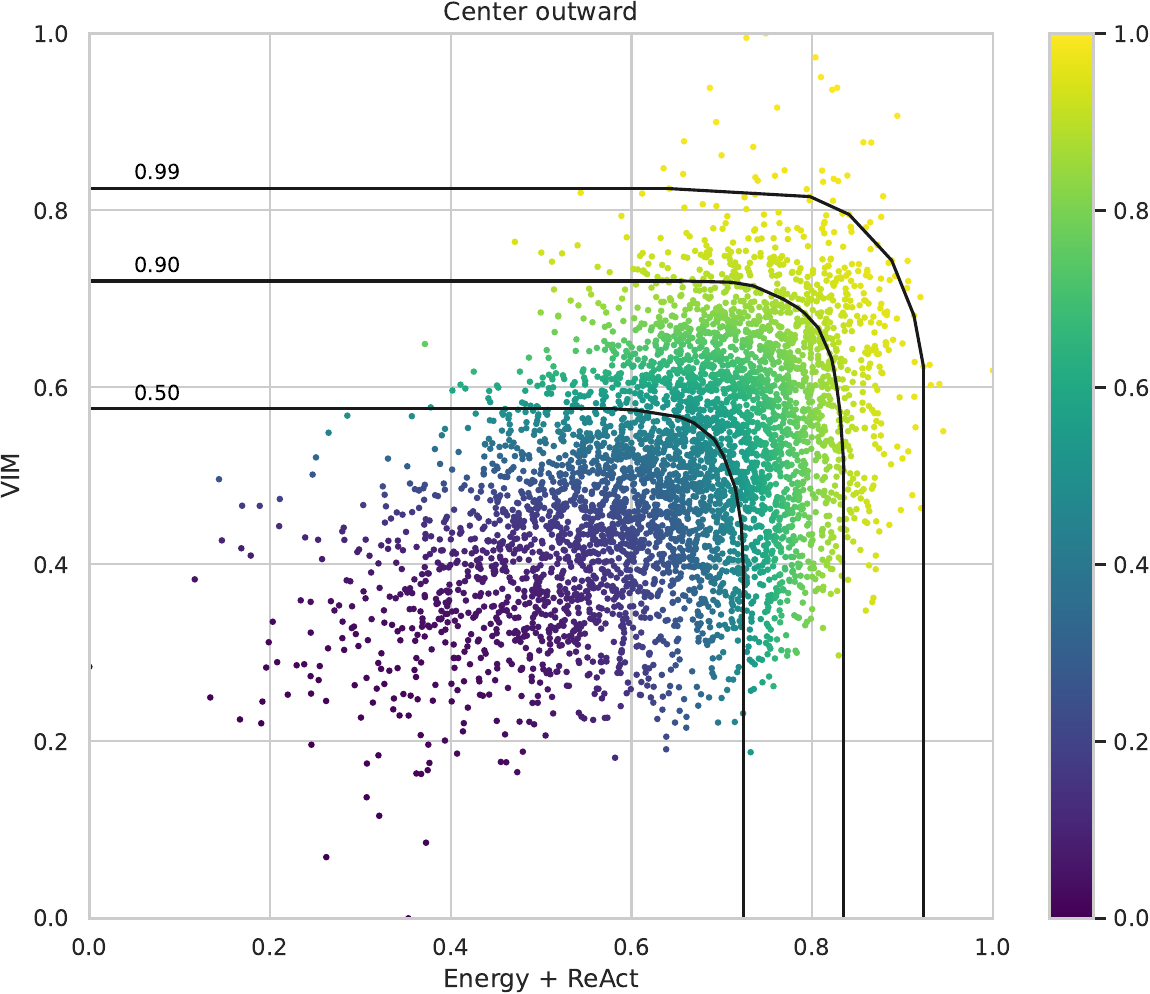}
    \caption{Visualization of sets $A_t$ for different combination methods with ImageNet-200 as ID data.\label{fig:quantileplot}}
    \vspace{-0.5cm}
  \end{figure*}

It is not straightforward to generalize the commonly used metrics for OOD detection
evaluation to the case of a \multioodscore{} $\vs$. Indeed,
the \multidetector{}s constructed above involve drawing a decision boundary 
(determined by the subset $A$) rather than simply using a threshold $\tau\in\R$.
In order to generalize the common metrics to the newly defined \multiooddetector{}s,
we choose to define a one-parameter family of sets $(A_t)_{0\leq t\leq 1}$ adapted to each method.
\begin{itemize}
    \item For the majority vote, $A_t$ is obtained by 
    picking thresholds $\tau_i$ such that the FPR associated to the individual score $s_i$ is equal to $t$.
    \item For the CDF based \multidetector{}s, $A_t$ can be defined by using iso-probability levels, e.g. for the empirical CDF $A_t=\big\lbrace \vy\in \R^d \,\big\rvert\, \hat{F}_{\vs(\rvx)}(\vy) > t \big\rbrace$.
    \item For the center-outward quantiles we take extended version of
    $A_t=\R^d_+ \setminus \text{Conv}\lbrace \vs^{(j)}\,|\, q_j \leq t\rbrace$.
    
\end{itemize}
This one-dimensional parametrization of the different decision boundaries allow
for the generalization of the AUROC as well as FPR$(t)$ and TPR$(t)$ metrics. We provide a visualization of such $A_t$ for different combination methods in Figure \ref{fig:quantileplot}.

\subsection{Search Strategies}

\vspace{-5pt}
The purpose of this section is to choose the best score combinations 
for a given task. We are using 28 individual scores as candidates for being
combined with each other. Since evaluating all the possible combinations
is computationally infeasible, we propose a few more lightweight alternatives.
The comparison between the different combinations is performed using both
ID and OOD data. The ID data that we use is that of the set $\smash{\setidval}$ (see Section~\ref{sec:background-ood}). 
Given an OOD dataset $\smash{\setood}$ we split it into two disjoint subsets,
the set $\smash{\setoodval}$ 
is used along with $\smash{\setidval}$ for model selection purposes,
while the set $\smash{\setoodtest}$ 
is used along the set $\smash{\setidtest}$ for final evaluation
purposes. The test sets are never used for score fitting, calibration of model combination, or model selection.

\subsubsection{Best Pairs}
A first approach to finding promising sets of scores to combine is to consider only pairs of individual score functions and simply test all the combinations on the validation datasets. In our case, we consider 28 different univariate OOD scores, which involves testing 378 combinations. Each of the individual OOD scores is only computed once, and can then be used in all of the combinations it is involved in. 
As a result, once the individual OOD scores are computed, such a benchmark can be performed in a few hours with a modern CPU.

\subsubsection{Sensitivity Analysis}
The previous method only allows to select pairs of univariate OOD scores. This section provides a methodology for selecting sets of more than two OOD scores using basic linear Sensitivity Analysis. Specifically, we sample $N=1000$ $d$-dimensional random binary vectors $\{\vb_1,..., \vb_N\}$ -- one dimension for each univariate OOD score --, and create a set of OOD scores to combine out of each of these vectors (the score is included for $1$'s and not included for $0$'s). We observed that combining more than 4 OOD scores seemed not to improve the AUROC, or even to decrease it, so we only drew sets of $4$ scores at most. 
We compute the AUROC of each of the $N$ \multiooddetector{} and store the result in
$\smash{\{z_1,\dots, z_N\}}$.
The AUROC values are then used to build an output variable 
$y_i := \1(z_i > z^p)$, where $z^p$ is the $p$-percentile of $\{z_1, ..., z_N\}$. 
Then, we perform a logistic regression of the pairs 
$\{ (\vb_1, y_1), ..., (\vb_N, y_N)\}$ and consider the coefficients of this regression as sensitivity indices. Note that using $y_i = \1(z_i > z^p)$ rather than $z_i$ allows us to assess the importance of each univariate score in reaching the $p$-percentile.  

\subsubsection{Beam Search}
Beam search, originally introduced in the context of speech recognition systems \cite{lowerre1976harpy}, is a heuristic algorithm that efficiently explores large search spaces by maintaining a fixed number of promising solutions at each step. This method strikes a balance between breadth and depth, enabling comprehensive exploration without exhaustive computation.

In our approach, we use beam search to identify optimal combinations of univariate OOD scores. The algorithm iteratively combines the best-performing scores, as evaluated on the calibration datasets, to form sets. At each iteration, we retain a specified number of top-performing combinations (beam width) and expand them by adding new scores, up to a predetermined maximum ensemble size (beam depth). This process allows us to systematically explore and identify effective OOD score combinations. For more details, refer to the pseudo-code of the algorithm in the Appendix \ref{app:comb}.

\vspace{-5pt}
\section{Experiments}\label{sec:experiments}

\vspace{-5pt}
In this section, we explore the performances of the combination methods and the search strategies on OpenOOD \cite{yang2022openood}, an extensive benchmark for OOD detection. We study $28$ individual OOD detectors listed in Appendix \ref{app:full_results}, using the Post-Hoc OOD detection library Oodeel \cite{oodeel}. We test the methods on three ID datasets, CIFAR-10, CIFAR-100 and Imagenet-200, with a set of OOD benchmarks gathered under two categories: Near-OOD and Far-OOD. The models considered are ResNet18 for CIFAR and ResNet50 for Imagenet200, trained on ID datasets, downloaded from OpenOOD. Note that the OOD datasets included in these two categories are not the same for each ID dataset. More details on the benchmark are found in Appendix \ref{app:openood}. In addition, if there are any, we defer implementation details of the combination methods and the search strategies to Appendix \ref{app:comb}.

We consider two approaches corresponding to common practical settings. The first case is when, as a practitioner, one has to choose between a large number of available OOD detectors and test them on their own data. In this case, we assume that some OOD data is available to select the most appropriate detector. In the second scenario, we consider that no OOD data is available for selecting a set of OOD detectors to combine. In that case, we draw inspiration from Outlier Exposure principles \cite{hendrycks2018deep} and select some external data as proxies for OOD data. We use data specifically curated as Outlier Exposure data by OpenOOD \cite{yang2022openood}. Details on these datasets are found in Appendix \ref{app:openood}.

\begin{table*}[ht!]
\centering
\resizebox{\textwidth}{!}{
\begin{tabular}{l|cc|cc|cc} 
    \toprule
     & \multicolumn{2}{c|}{\textbf{CIFAR-10}} & \multicolumn{2}{c|}{\textbf{CIFAR-100}} & \multicolumn{2}{c}{\textbf{ImageNet-200}} \\
    
    \midrule
         & {Near OOD} & {Far OOD}  & {Near OOD} & {Far OOD}  & {Near OOD} & {Far OOD}  \\
    \midrule
    Best indiv.               & {90.7} & {93.3} & {80.4}   & {86.6}  & {83.8}  & {93.0}  \\ 
    \midrule
    \multicolumn{7}{l}{\textbf{Majority Vote} (loose)} \\
    \midrule
    Best pair                  & {90.3} \gain{-0.4} & {96.7} \gain{3.4} & {80.6} \gain{0.2} & {88.8} \gain{2.2} & {83.8} \gain{0.0} & {94.2} \gain{1.2}  \\
    Sensitivity                & \textbf{91.1 \gain{0.4}} & {95.6} \gain{2.3} & {81.9} \gain{1.5} & {88.8} \gain{2.2} & {87.1} \gain{3.3} & {94.2} \gain{1.2}  \\ 
    Beam Search                & 90.7 \gain{0.0} & 95.6 \gain{2.4} & 80.6 \gain{0.3} & 89.1 \gain{2.5} & 84.5 \gain{0.7} & 94.8 \gain{1.8} \\ 
    \midrule
    \multicolumn{7}{l}{\textbf{Empirical CDF}} \\
    \midrule
    Best pair                  & {90.2} \gain{-0.5} & {96.1} \gain{2.8} & {80.7} \gain{0.3} & {88.5} \gain{1.9} & {85.5} \gain{1.7} & {95.1} \gain{2.1}  \\ 
    Sensitivity                & {90.2} \gain{-0.5} & {96.1} \gain{2.8} & \textbf{82.2 \gain{1.8}}  & {88.0} \gain{1.4}  & \textbf{88.3 \gain{4.5}}  & {95.1} \gain{2.1}  \\ 
    Beam Search                & 90.7 \gain{0.0} & 96.1 \gain{2.8} & 80.6 \gain{0.2} & 89.3 \gain{2.7}  & 85.6 \gain{1.7} & 95.3 \gain{2.3} \\ 
    \midrule
    \multicolumn{7}{l}{\textbf{Copulas} }\\
    \midrule
    Best pair                & {90.7} \gain{0.0} & {96.7} \gain{3.4} & {80.3} \gain{-0.1} & {88.7} \gain{2.1}  & {85.1} \gain{1.9} & {95.2} \gain{2.2}  \\ 
    Sensitivity                &  {90.7} \gain{0.0} & {96.7} \gain{3.4} & {81.8} \gain{1.4} & {88.4} \gain{1.8} & {88.2} \gain{4.5} & {95.1} \gain{2.1} \\ 
    Beam Search                & {90.7} \gain{0.0} & \textbf{96.8 \gain{3.5}} & 80.6 \gain{0.2} & \textbf{90.7 \gain{4.1}} & 85.1 \gain{1.3} & \textbf{95.8 \gain{2.8}} \\ 
    \midrule
    \multicolumn{7}{l}{\textbf{Center Outward}  (knn)} \\
    \midrule
    Best pair                & {90.4} \gain{-0.3} & {96.4} \gain{3.1} & {80.6} \gain{0.2} & {89.3} \gain{2.7} & {84.7} \gain{0.9} & {95.3} \gain{2.3} \\ 
    Sensitivity                &  {90.9} \gain{0.2} & {96.5} \gain{3.2} & {81.3} \gain{0.9}  & {89.0} \gain{2.4} & \textbf{88.3 \gain{4.5}} & {94.5} \gain{1.5} \\ 
    Beam Search                & {90.7} \gain{0.0} & 96.4 \gain{3.1} & 80.9 \gain{0.5} & 89.9 \gain{3.3} & {85.2} \gain{1.4} & 95.3 \gain{2.3} \\ 
    \bottomrule
    \end{tabular}
}
\caption{AUROC of combination methods for best performing sets of individual OOD detectors selected with different search strategies.Best pair - selected as the best pair on the validation dataset; Sensitivity OE - selected after a sensitivity analysis performed on a validation dataset;  Beam Search - Selected after a Beam Search on a validation dataset}
\label{tab:main_results}
\end{table*}

\subsection{Combination Methods and Search Strategies on OpenOOD}

\begin{wraptable}{r}{0.6\textwidth}
\vspace{-0.4cm}
    \centering
    \resizebox{0.6\textwidth}{!}{
    \begin{tabular}{l|l|p{7cm}|l}
        \toprule
        \textbf{Dataset} & \textbf{Experiment} & \textbf{Best Tuple} & \textbf{Best Indiv.} \\ \midrule
        \multirow{2}{*}{CIFAR-10}   & Near-OOD   & DKNN, Energy & DKNN\\ 
                                    & Far-OOD    & DKNN, Gram, VIM & Gram \\ \midrule
        \multirow{2}{*}{CIFAR-100}  & Near-OOD   & Entropy, GEN, RMDS, VIM & GEN\\
                                    & Far-OOD    & DKNN, Gram, ODIN + ReAct, RMDS & Gram\\ \midrule
        \multirow{2}{*}{ImageNet-200} & Near-OOD & DKNN, Entropy, MLS + ASH, ODIN + ASH & Entropy\\
                                      & Far-OOD  & Energy + ASH, GEN + ReAct, MLS + SCALE, VIM & Energy + SCALE\\ 
        \bottomrule
    \end{tabular}}
    \caption{Best tuples for OOD detection methods using Copula, selected via beam search. This table highlights the diversity of the retained methods across different experiments.}
    \vspace{-0.2cm}
    \label{tab:beam_seach_tuples_examples}
\end{wraptable}

In this part, we present the results of each of the four combination methods, with combination sets selected using the different search strategies. For each set of univariate OOD detectors to combine, the \multidetector{} is fit on $\setidcal$. The search strategies consider AUROCS for each \multidetector{} on $\setidval$ and $\setoodval$. Once the best set is selected, it is tested on $\setidtest$ $\setoodtest$ for comparison on OpenOOD with the univariate detectors. The split used is $25-25-50\%$ of ID OpenOOD test data for $\setidcal$, $\setidval$ and $\setidtest$, and $50-50\%$ of OOD OpenOOD test data for $\setoodval$ and $\setoodtest$ respectively -- recall that we don't need OOD calibration data since \multidetector{}s are only fit on ID data.

The results are gathered in Table \ref{tab:main_results}. 
We only report the best individual method; the detailed results for all individual scores are deferred to Appendix \ref{app:full_results}. 
The results are quite satisfying: every search strategy finds a set of scores to combine that at least improves the AUROC of the best-performing individual detector on every benchmark. 
For Majority Vote and Center Outward, the Sensitivity method always improves the best individual detector, and some combination exhibits quite significant gains, such as $3.5\%$ for CIFAR-10 Far-OOD, $4.1\%$ for CIFAR-100 Far-OOD or $4.5\%$ for ImageNet-200 Near-OOD (which is the most challenging of the three benchmarks), with several other improvements of more than $1\%$.  
We report the best tuples found by the Beam Search strategy for each test case in Table \ref{tab:beam_seach_tuples_examples}. 
The diversity of the optimal tuples illustrates the necessity of an elaborated search strategy.

\vspace{-0.4cm}

\subsection{Selecting Promising Sets Without OOD Data}\label{sec:oe} 

\begin{wrapfigure}{l}{0.6\textwidth}
\vspace{-0.5cm}
  \centering
  \includegraphics[width=0.6\textwidth]{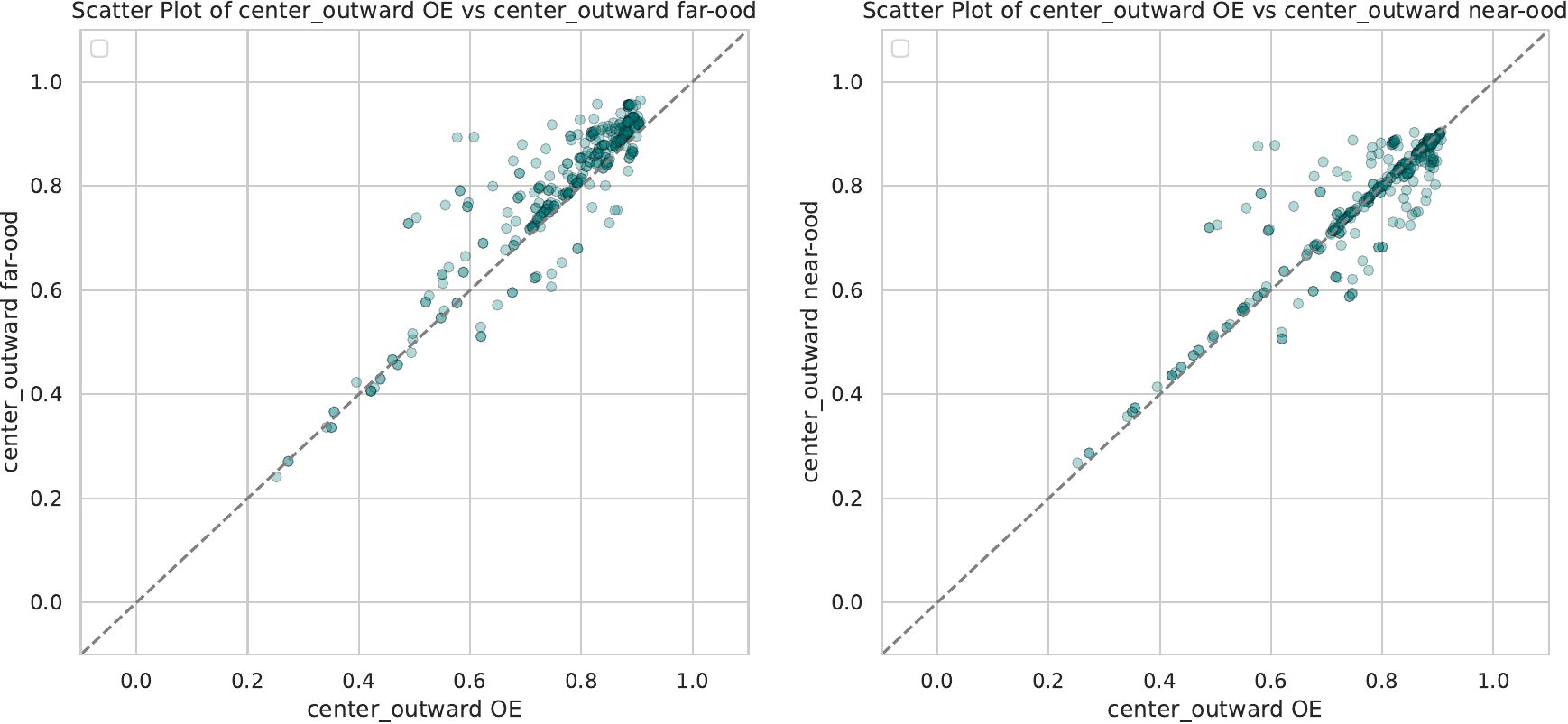}
    \caption{Scatter plot of AUROC obtained for pairs of OOD detectors combined with Center Outward method x-axis: on proxy OOD data (OE) and y-axis: on near OOD and far OOD OpenOOD benchmark for Cifar10 considered in-distribution. \label{fig:corr_oe}}
    \vspace{-0.5cm}
\end{wrapfigure}

In many real-world cases, the access to OOD data is limited. Outlier Exposure \cite{hendrycks2018deep} has emerged to cope with that lack of OOD data. It is now a common practice to use Outlier Exposure (OE), i.e. selecting external datasets as proxies for OOD data, to tune OOD detectors \cite{dong2022neural,wang2022few,wang2022out,wang2022watermarking,meinke2021provably,huang2022density,fort2021exploring}. OpenOOD benchmark actually curates datasets labeled Outlier Exposure for each ID dataset, which is strictly disjoint from the corresponding test datasets. In this section, based on the intuition that combinations of detectors that are performant on Outlier Exposure might be performant on test data -- as illustrated in Figure \ref{fig:corr_oe} --, we investigate the use of Outlier Exposure to select promising sets of OOD detectors.

We fit the \multiooddetector{}s 
on the same $\smash{\setidfit}$ dataset, but perform the set selection with the search strategies based on AUROC computed using $\smash{\setidval}$ and a new $\smash{\mathcal{D}_{oe}^{val}}$ constructed using Outlier Exposure from the datasets curated by OpenOOD. For the Beam Search, we return the top $9$ sets (the best of each level with a width of 3 and depth of 4),
for the sensitivity analysis, the top $11$ (the total number of combinations from the set of 4 best detectors identified with the sensitivity analysis) and for the best pair, the top 18 sets (Top $5\%$ of tested pairs). We defer the full results in the form of a Table and Pareto fronts to Appendix \ref{app:oe_results}, 
but for illustration purposes, we pick one Pareto plot for the Copula method with Sensitivity strategy that represents test AUROCs of the returned sets for near OOD as x-axis and far OOD as y-axis. This plot shows that many of these sets are outside the Pareto front of individual detectors. Such improvements are also observed for other \multidetector{}s and search strategies, as shown in the Appendix \ref{app:oe_results}.

\begin{figure*}[!h]
  \label{fig:pareto}
  \centering
  \includegraphics[width=1.0\textwidth]{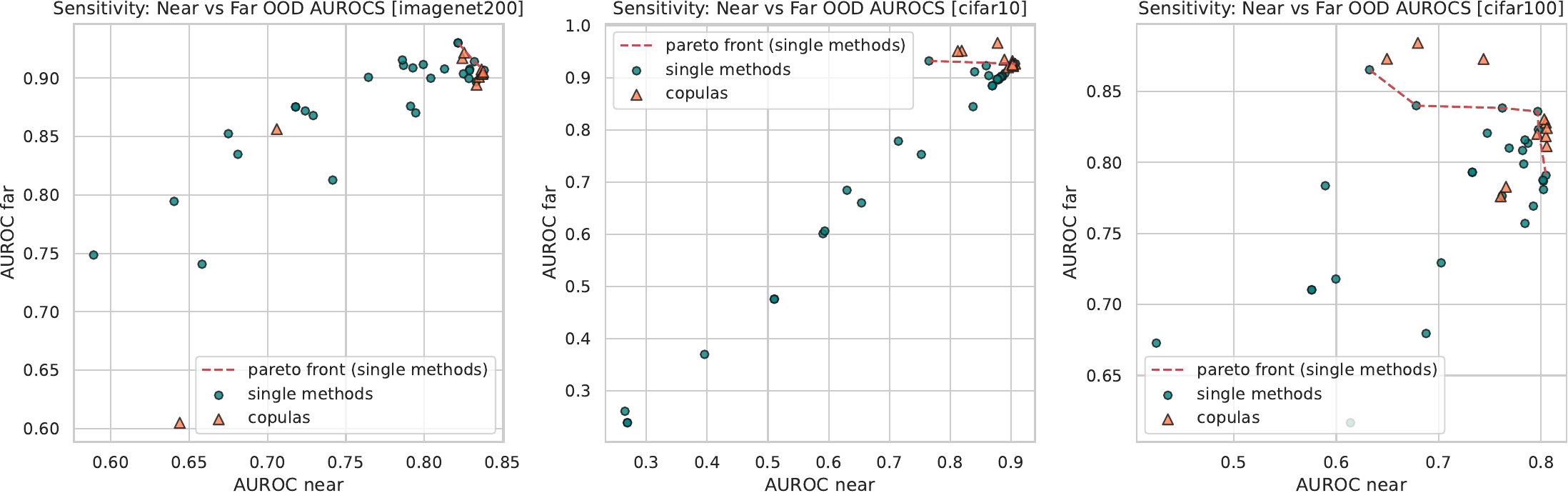}
    \caption{Pareto fronts with individual methods and sets of detectors combined with Copulas returned after a Sensitivity Analysis based on AUROCS obtained on Outlier Exposure datasets.}
  \end{figure*}

\section{Limitations}\label{sec:limitations}
While we believe that combining OOD scores is good way to boost the performance of individual OOD detectors, 
we acknowledge that the methods proposed have limitations:
\textbf{ID data availability.} Our methods involve using a training set, a calibration set, a validation set and a test set, which might be prohibitive for applications in which data is scarce. \textbf{OOD data availability.} Picking the best combination of scores requires OOD data (proxy or not), while we believe that this might be a limitation in some context like Anomaly Detection where OOD data is rather scarce, we are also aware that using OOD data is a common practice for model selection and benchmarking purposes. \textbf{Extra compute resources.} We are aware that searching for the best combination of scores carries an extra computation effort compared
to selecting the best individual scores. However, combinations of over four scores provide negligible improvement, and some OOD detector can be implemented jointly to avoid repeated forward passes. We thus believe that the extra compute power required is only a minor limitation of our method.

\section{Conclusion}

In this work, we investigated the combination of existing OOD detectors to improve OOD detection capabilities. We introduced four different combination methods based on majority vote, empirical CDF, parametric CDF with copulas or center outward. We also introduced several search strategies to choose which individual detectors to combine and showed that this whole methodology could significantly improve the state-of-the-art for CIFAR-10, CIFAR-100, and ImageNet-200 test cases as designed on the OpenOOD benchmark. The methodology exhibits several additional advantages. It is versatile because it can be applied to any reference metrics -- we use AUROC, but depending on the industrial application, it is possible to apply it to any other metric, e.g., FPR@TPR95 or TPR@FPR5. Finally, it will stay automatically up to date with state-of-the-art post-hoc OOD detection because its performance stems from the underlying individual OOD detectors. 

\bibliographystyle{unsrt}
\bibliography{refs}

\appendix
\newpage

\section{Full results for individual OOD detectors}\label{app:full_results}

\begin{table*}[!h]
\centering
\resizebox{\textwidth}{!}{%
\begin{tabular}{l|cc|cc|cc}
\toprule
OOD type & \multicolumn{2}{c}{\textbf{CIFAR-10}} & \multicolumn{2}{c}{\textbf{CIFAR-100}} & \multicolumn{2}{c}{\textbf{ImageNet-200}} \\
\midrule
& Near OOD & Far OOD & Near OOD & Far OOD & Near OOD & Far OOD \\
\midrule
Entropy\cite{ren2019likelihood} & 88.49 & 90.33 & 80.23 & 78.09 & \textbf{83.83} & 90.67 \\
GEN\cite{liu2023gen} & 89.03 & 90.86 & \textbf{80.47} & 79.09 & 82.90 & 90.77 \\
MLS\cite{vaze_open-set_2022}& 87.74 & 89.70 & 80.23 & 78.68 & 82.88 & 90.64 \\
MSP\cite{hendrycks2016baseline} & 88.26 & 90.07 & 79.24 & 76.92 & 83.31 & 89.69 \\
Energy\cite{liu_energy-based_2021}& 87.80 & 89.75 & 80.17 & 78.75 & 82.50 & 90.37 \\
RMDS\cite{ren2021simple} & 85.93 & 92.28 & 79.67 & 83.59 & 82.85 & 89.97 \\
GEN\cite{liu2023gen} + SCALE\cite{xu2023scaling} & 75.25 & 75.30 & 76.19 & 77.64 & 83.20 & 91.41 \\
DKNN\cite{sun_out--distribution_2022} & \textbf{90.75} & 92.66 & 79.73 & 82.31 & 81.29 & 90.78 \\
ODIN\cite{liang_enhancing_2020} & 71.47 & 77.83 & 76.21 & 83.83 & 79.27 & 90.87 \\
MSP\cite{hendrycks2016baseline} + ReAct\cite{react21nips} & 87.86 & 89.56 & 78.30 & 79.89 & 80.42 & 89.99 \\
MLS\cite{vaze_open-set_2022} + ReAct\cite{react21nips} & 86.91 & 88.44 & 78.19 & 80.84 & 79.94 & 91.16 \\
MLS\cite{vaze_open-set_2022} + SCALE\cite{xu2023scaling} & 51.07 & 47.61 & 73.28 & 79.30 & 82.16 & 93.01 \\
Energy\cite{liu_energy-based_2021} + SCALE\cite{xu2023scaling} & 51.03 & 47.57 & 73.28 & 79.32 & 82.15 & \textbf{93.02} \\
Entropy\cite{ren2019likelihood} + ReAct\cite{react21nips} & 88.05 & 89.73 & 78.72 & 81.35 & 78.68 & 91.09 \\
SHE\cite{zhang2022out} & 83.76 & 84.45 & 78.42 & 75.70 & 79.45 & 87.02 \\
VIM\cite{wang2022vim} & 86.35 & 90.39 & 74.74 & 82.06 & 78.60 & 91.55 \\
GEN\cite{liu2023gen} + ASH\cite{djurisic2022extremely} & 59.35 & 60.63 & 70.24 & 72.93 & 79.13 & 87.60 \\
GEN\cite{liu2023gen} + ReAct\cite{react21nips} & 88.61 & 90.22 & 78.42 & 81.58 & 76.44 & 90.07 \\
MSP\cite{hendrycks2016baseline} + SCALE\cite{xu2023scaling} & 65.40 & 66.01 & 68.77 & 67.95 & 74.16 & 81.27 \\
ODIN\cite{liang_enhancing_2020} + ReAct\cite{react21nips} & 63.01 & 68.44 & 67.81 & 84.00 & 67.50 & 85.24 \\
ODIN\cite{liang_enhancing_2020} + SCALE\cite{xu2023scaling} & 39.57 & 37.01 & 58.93 & 78.36 & 72.91 & 86.80 \\
Energy\cite{liu_energy-based_2021} + ReAct\cite{react21nips} & 86.97 & 88.48 & 76.88 & 81.01 & 72.40 & 87.19 \\
MLS\cite{vaze_open-set_2022} + ASH\cite{djurisic2022extremely} & 26.88 & 23.93 & 57.61 & 71.03 & 71.79 & 87.54 \\
Energy\cite{liu_energy-based_2021} + ASH\cite{djurisic2022extremely} & 26.86 & 23.91 & 57.59 & 71.03 & 71.78 & 87.53 \\
MSP\cite{hendrycks2016baseline} + ASH\cite{djurisic2022extremely} & 59.06 & 60.12 & 61.38 & 61.67 & 65.81 & 74.08 \\
Gram\cite{gram20icml} & 76.51 & \textbf{93.20} & 63.24 & \textbf{86.52} & 68.10 & 83.48 \\
Mahalanobis\cite{lee2018simple} & 84.02 & 91.12 & 59.96 & 71.79 & 64.03 & 79.45 \\
ODIN\cite{liang_enhancing_2020} + ASH\cite{djurisic2022extremely} & 26.47 & 26.11 & 42.43 & 67.28 & 58.89 & 74.86 \\

\bottomrule
\end{tabular}%
}
\caption{AUROC scores for different OOD detection methods across various datasets.}
\end{table*}

\section{Additional Details on Combination Methods and Search Strategies}\label{app:comb}

This section presents more details about the combination methods and search strategies. We also specify some implementation details when appropriate.

To begin with, the OOD scores from individual methods have all been obtained using the post-hoc OOD detection library Oodeel \cite{oodeel}. For each ID dataset (Cifar10, Cifar100, and Imagenet200), the computation has been conducted on an Nvidia RTX 4090 or RTX 3080 GPU. Once the individual scores have been stored, all the experiments we perform with them can be carried out on a standard laptop CPU.

\subsection{Majority vote}

Compared to other combination methods, the majority vote approach does not combine scores but rather combines binary decisions (ID or OOD). For each individual OOD detector, thresholds $\tau_i$ are used to make the decision.

There are several ways to combine multiple binary values and produce a single binary decision. We defined and experimented with four of these methods:
\begin{itemize}
    \item \textbf{ALL}: An element is declared OOD if all individual OOD detectors classify it as OOD.
    \item \textbf{ANY}: An element is declared OOD if at least one individual OOD detector classifies it as OOD.
    \item \textbf{LOOSE}: An element is declared OOD if at least half of the individual OOD detectors classify it as OOD (with a tie-breaking rule in favor of OOD).
    \item \textbf{STRICT}: An element is declared OOD if more than half of the individual OOD detectors classify it as OOD (with a tie-breaking rule in favor of ID).
\end{itemize}

Note that the STRICT and LOOSE majority vote methods are equivalent when the number $d$ of individual methods is odd. Additionally, when $d=2$, the ALL and STRICT methods are equivalent, as are the ANY and LOOSE methods. Our experimental findings indicate that the STRICT and LOOSE approaches are the most effective, with a slight preference for the LOOSE method.

\subsection{Copulas}

As stated in the main paper, the copula method involves choosing a parametric distribution for marginal distributions, and a copula function, which can be chosen among different types of parametric copulas. 

We selected the marginal distribution among Gaussian, Beta, and Uniform distributions and found that choosing Uniform consistently outperformed other distributions. As for the Copulas, we considered 6 possible different functions parametrized by $\theta$:

\paragraph{Clayton Copula (bivariate)}

The Clayton Copula function is defined as:
$$
C(u_1, u_2) = (u_1^{-\theta} + u_2^{-\theta} - 1)^{-1/\theta}
$$
where $u_1, u_2 \in [0, 1]^2$.

\paragraph{Frank Copula (bivariate)}

The Clayton Copula function is defined as:
$$
C(u_1, u_2) = -\frac{1}{\theta}
              log \left( 1 +
                         \frac{(e^{-\theta u_1} - 1)(e^{-\theta u_2} - 1)}
                              {e^{-\theta} - 1}
                  \right)
$$
where $u_1, u_2 \in [0, 1]^2$.

\paragraph{Gumbel Copula (bivariate)}

The Clayton Copula function is defined as:

\begin{eqnarray*}
  \displaystyle \Hat{\theta}_n=\frac{1^{\strut}}{1 - \Hat{\tau}_{n_{\strut}}}
\end{eqnarray*}

where $u_1, u_2 \in [0, 1]^2$, and $\Hat{\tau}_{n_{\strut}}$ is the Kendall-$\tau$, which measures the association between pairs of data points. It is defined as:

\[ \tau = \frac{C - D}{\frac{1}{2}n(n-1)} \]

where:
- \( C \) is the number of concordant pairs.
- \( D \) is the number of discordant pairs.
- \( n \) is the number of data points.

For a given set of \( n \) pairs \((u^{(i)}_1, u^{(i)}_2)\) where \( i = 1, 2, \ldots, n \):

\begin{itemize}
    \item \textbf{Concordant Pair}: A pair of observations \((u^{(i)}_1, u^{(i)}_2)\) and \((u^{(j)}_1, u^{(j)}_2)\) is concordant if the order of \( u^{(i)}_1 \) and \( u^{(j)}_1 \) is the same as the order of \( u^{(i)}_2 \) and \( u^{(j)}_2 \). That is, either both \( u^{(i)}_1 < u^{(j)}_1 \) and \( u^{(i)}_2 < u^{(j)}_2 \) or both \( u^{(i)}_1 > u^{(j)}_1 \) and \( u^{(i)}_2 > u^{(j)}_2 \).
    \item \textbf{Discordant Pair}: A pair of observations \((u^{(i)}_1, u^{(i)}_2)\) and \((u^{(j)}_1, u^{(j)}_2)\) is discordant if the order of \( u^{(i)}_1 \) and \( u^{(j)}_1 \) is different from the order of \( u^{(i)}_2 \) and \( u^{(j)}_2 \). That is, either \( u^{(i)}_1 < u^{(j)}_1 \) and \( u^{(i)}_2 > u^{(j)}_2 \) or \( u^{(i)}_1 > u^{(j)}_1 \) and \( u^{(i)}_2 < u^{(j)}_2 \).
\end{itemize}

\paragraph{Plackett Copula (bivariate)}

The Clayton Copula function is defined as:
$$
C(u_1, u_2) = \frac{\left[1+(\theta-1)(u_1+u_2)\right]-
              \sqrt{\left[1+(\theta-1)(u_1+u_2)\right]^2-
              4u_1u_2\theta(\theta-1)}}{2(\theta-1)}
$$
where $u_1, u_2 \in [0, 1]^2$.

\paragraph{Normal Copula (multivariate)}

The Normal copula is defined by :

$$C(u_1, \cdots, u_n) = \Phi_{\mR}^n(\Phi^{-1}(u_1), \cdots, \Phi^{-1}(u_n))$$

where $\Phi_{\mR}^n$ is the cumulative distribution function of the normal distribution with zero mean, unit marginal variances and correlation R:

$$\Phi_{\mR}^n(\vx) = \int_{-\infty}^{x_1} \ldots
                               \int_{-\infty}^{x_n}
                               \frac{1}
                                    {{(2\pi\operatorname{det}\mR)}^{\frac{n}{2}}}
                             \exp \left(-\frac{\vu^T \mR \vu}{2} \right)d\vu
$$
with $\Phi$ given by:

$$\Phi(x) = \int_{-\infty}^x \frac{1}{\sqrt{2\pi}} e^{-\frac{t^2}{2}}dt$$

The correlation matrix $\mR$ is linked to the Spearman correlation and the Kendall concordance through the following relations:

From the Spearman correlation matrix:

$$\mR_{ij} = 2 \sin \left( \frac{\pi}{6}\rho_{ij}^S \right)$$

where $\rho_{ij}^S = \rho^S(X_i,X_j) = \rho^P(F_{X_i}(X_i),F_{X_j}(X_j))$

From the Kendall concordance matrix:

$$\mR_{ij} = \sin \left( \frac{\pi}{2} \tau_{ij} \right)$$

with

$$\tau_{ij} = \tau(X_i,X_j)
          = \mathbb{P}{(X_{i_1} - X_{i_2})(X_{j_1} - X_{j_2}) > 0} -
            \mathbb{P}{(X_{i_1} - X_{i_2})(X_{j_1} - X_{j_2}) < 0}$$

where $(X_{i_1},X_{j_1}$ and $(X_{i_2},X_{j_2})$ follow the distribution of $(X_i,X_j)$.

\paragraph{Independant Copula (multivariate)}

The Independent Copula function is defined as:
$$
C(u_1, \cdots, u_n) = \prod_{i=1}^n u_i
$$
where $u_i \in [0, 1]$.

For the bivariate case, i.e., experiments involving only pairs, we selected Frank Copula. Surprisingly, the Independent Copula outperformed the others in the multivariate case (found in the Sensitivity and Beam search strategies).

We used the library Openturns \cite{Baudin2016}, which implements many distributions and copulas families, as well as maximum likelihood estimation, to fit their parameters to data. 

Some elements of the presentation of this appendix are extracted from openturn's documentation \cite{Baudin2016}.

\subsection{Center Outward}

\subsubsection{Monotonicity property on convex hulls}

\begin{figure}[h!]
    \centering
    \includegraphics[width=\textwidth]{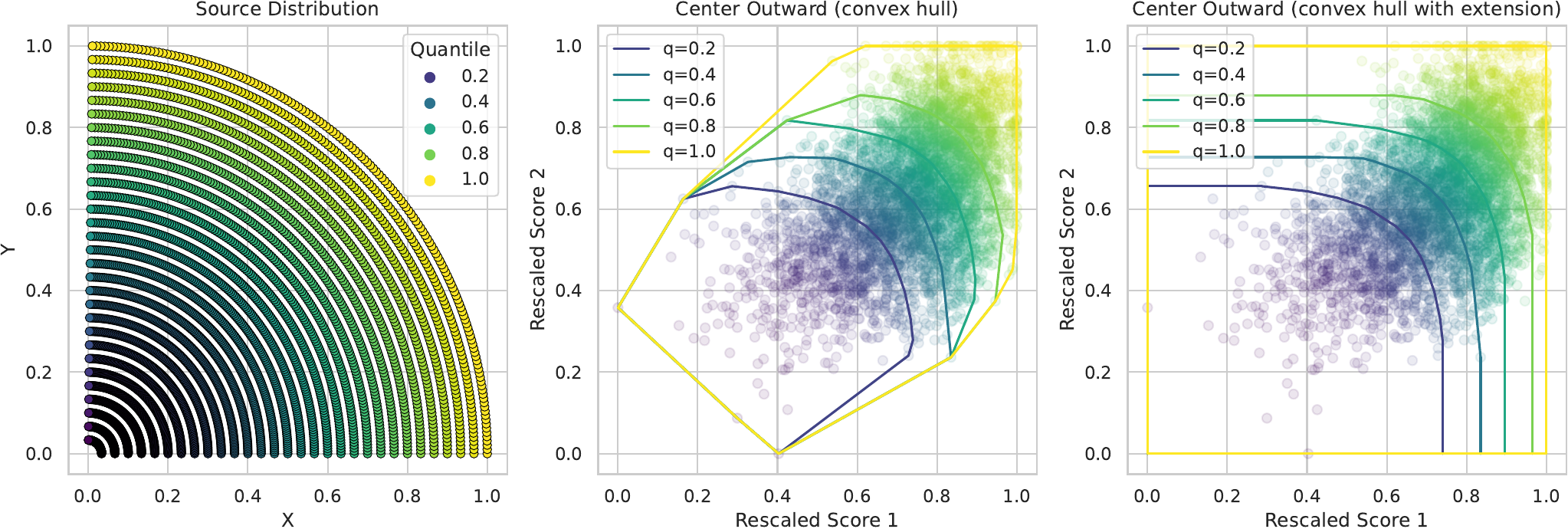}
    \caption{\textbf{(Left)} Source distribution: nested hyperspheres intersected with quadrant $\R^2_+$}; \textbf{(Middle)} Convex hulls for 5 different quantile values; \textbf{(Right)} Extended convex hulls (ensuring monotonicity property) for 5 different quantile values
    \label{fig:center_outward_monotony}
\end{figure}

In order to ensure monotonicity property as defined in~\ref{sec:baselines}, and claimed in~\ref{sec:co_ood_classif}, we need to extend the ID points contained in the convex hull so that if a point is in the hull, all points beneath it towards the origin or any axes are also included. This property is illustrated in Figure \ref{fig:center_outward_monotony} and the building process is as follows: For each dimension, the points included in the original hull are projected onto a plane by removing one dimension at a time. For each projected set of points, the algorithm attempts to construct a convex hull. If successful, it recursively ensures the monotony property for the projected hull points. Then, for each projected dimension, points aligned towards the origin are added back to the hull to ensure all points beneath any included point are also part of the hull. Finally, the original points and the newly added points are combined to form the monotone convex hull. 

\subsubsection{Generalization of quantiles: Convex hulls vs KNN}
\label{sec:knn_gen}

\begin{figure}[h]
    \centering
    \includegraphics[width=0.5\textwidth]{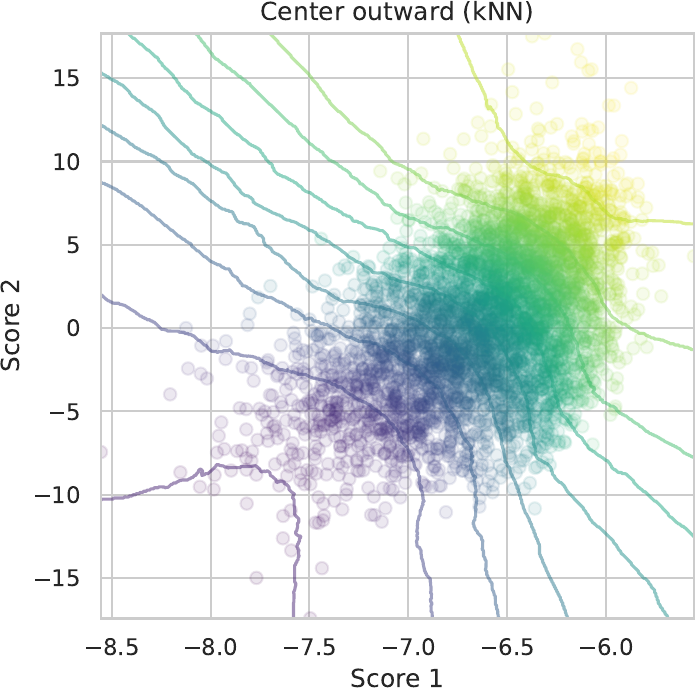}
    \caption{Generalization of center outward quantiles on $\R^2_+$ using KNN instead of convex hull.}
    \label{fig:center_outward_knn}
\end{figure}

Using convex hulls for center-outward quantiles ensures generalization for new points by finding the smallest convex hull that contains them. However, this process can become highly time-consuming as the number of scores, denoted by $d$, increases. Indeed, points are added by iteratively projecting in hyperplanes of lower dimensions, resulting in a complexity that scales super-exponentially. 
To address this issue, we propose a more efficient generalization approach using K-Nearest Neighbors (KNN). The principle is straightforward: when estimating quantiles for in-distribution calibration scores $\setidcal$, we fit a KNN model on these points. For a new point, we assign a quantile value based on the mean quantile of the k-nearest ID points. This method significantly reduces computational time. Figure~\ref{fig:center_outward_knn} illustrates the quantile estimation over $\R_+^2$.


\subsubsection{Center-outward Implementation details}

Center outward results obtained in the reported tables are fitted with the following tricks / parameters:  

\begin{itemize}
    \item Scaling of data points: We rescale multidimensional score points in $[0,\, 1]$ using a QuantileTransformer form Sklearn library, this is useful to be more robust to outliers and scale discrepancies between the different ood detectors.
    \item Generalization of quantile estimation: For the computational reasons evoked in~\ref{sec:knn_gen}, we use KNN approach with $k=5$ to generalize quantiles, instead of the convex hulls.
    \item Sinkhorn regularization: We compute optimal transport maps using sinkhorn algorithm with POT library, with a regularization parameter of $0.01$.
\end{itemize}

\subsection{Sensitivity Analysis}

To obtain indices from the logistic regression, we used the \texttt{LogisticRegression} class of scikit-learn \cite{pedregosa2011scikit}

\subsection{Beam Search}

\subsubsection{Pseudo-code}
The pseudo-code of Algorithm \ref{alg:beam-search} illustrates the beam search process, which depends on two parameters: the beam width \codevar{w} and the beam depth \codevar{d}. Note that AUROC score for a given tuple is computed by fitting the aggregation method on \codevar{id\_scores\_cal} restricted to the methods of the tuple, and by evaluating the aggregation method over \codevar{id\_scores\_val} and \codevar{ood\_scores\_val}.

\begin{algorithm}[h]
\caption{Beam Search for \multiooddetector{}.}
\begin{spacing}{1.3}
\begin{algorithmic}[1]
\State \textbf{Input:} ID scores \codevar{id\_scores\_cal}, \codevar{id\_scores\_val}, OOD scores \codevar{ood\_scores\_val}, beam width \codevar{w}, beam depth \codevar{d}
\State \textbf{Output:} Best \multiooddetector{}
\State \textbf{Initialize:} Compute AUROC for each detector, retain top \codevar{w} detectors
\State \codevar{prev\_tuples} $\gets$ Top \codevar{w} individual detectors {\color{pleasantgreen}\Comment{\codevar{w} tuples of length 1}}
\State \codevar{best\_overall} $\gets$ Best individual detector
\State \codevar{best\_score} $\gets$ AUROC of best individual detector
\For{\codevar{i = 2} to \codevar{d}} {\color{pleasantgreen}\Comment{depth iteration}}
    \State \codevar{new\_tuples} $\gets \emptyset$
    \For{each \codevar{tuple} in \codevar{prev\_tuples}}
        \State \codevar{detectors\_to\_explore} $\gets$ Detectors not in \codevar{tuple}
        \For{each \codevar{detector} in \codevar{detectors\_to\_explore}}
            \State \codevar{new\_tuple} $\gets$ \codevar{tuple} $\cup \{\codevar{method}\}$
            \State Compute AUROC for \codevar{new\_tuple}
            \If{AUROC of \codevar{new\_tuple} $>$ \codevar{best\_score}}
                \State \codevar{best\_overall} $\gets$ \codevar{new\_tuple}
                \State \codevar{best\_score} $\gets$ AUROC of \codevar{new\_tuple}
            \EndIf
            \State \codevar{new\_tuples} $\gets$ \codevar{new\_tuples} $\cup \{\codevar{new\_tuple}\}$
        \EndFor
    \EndFor
    \State \codevar{prev\_tuples} $\gets$ Top \codevar{w} tuples from \codevar{new\_tuples} based on AUROC 
\EndFor
\State \Return \codevar{best\_overall}
\end{algorithmic}
\end{spacing}
\end{algorithm}
\label{alg:beam-search}

\subsubsection{Impact of depth and width parameters on Beam Search}

\begin{figure}[h!]
    \centering
\includegraphics[width=0.5\textwidth]{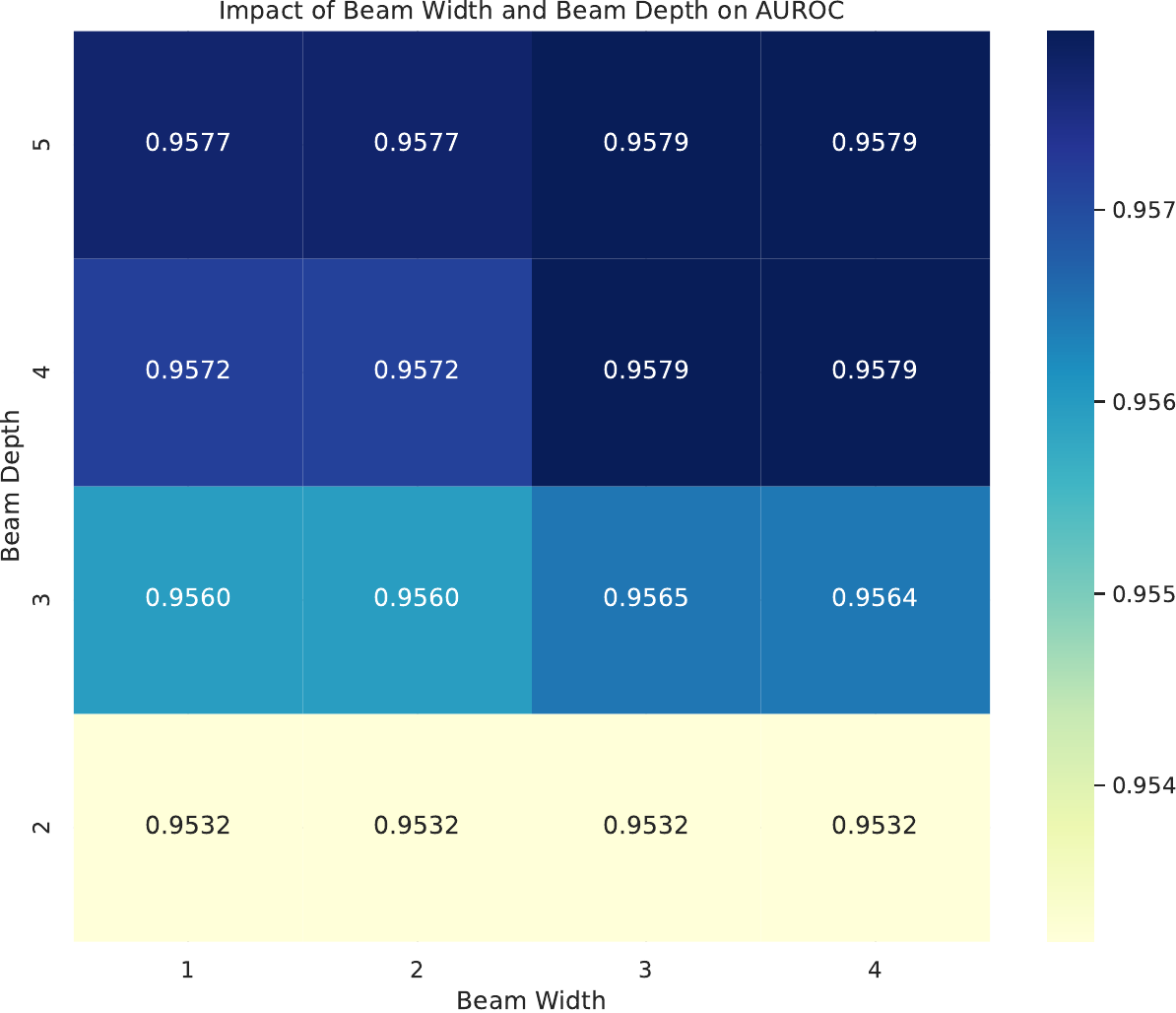}
    \caption{\textbf{Impact of beam search parameters on AUROC values.}}
    \label{fig:beam-search-parameters}
\end{figure}

The beam search algorithm relies on two critical parameters: the width and the depth of the search. To evaluate the significance of these parameters and identify an optimal configuration, we applied the algorithm with varying beam search depths and widths in the ImageNet200 vs. far-OOD experiment using the copulas approach. The results, presented in Figure \ref{fig:beam-search-parameters}, illustrate the AUROC values obtained. Based on the heatmap, a width of 3 and a depth of 4 provide a favorable balance between performance and computation time.

\section{Additional Plots for Outlier Exposure Motivations}\label{app:oe_plots}

In this section, we provide additional scatter plots of AUROC obtained for pairs of OOD detectors combined with combination methods, x-axis: on proxy OOD data (OE) and y-axis: on near OOD and far OOD OpenOOD benchmark for different datasets considered in-distribution.

\begin{figure*}[!h]
  \label{fig:ccpv}
  \centering
  \includegraphics[width=0.8\textwidth, trim={2cm, 1cm, 2cm, 1cm}]{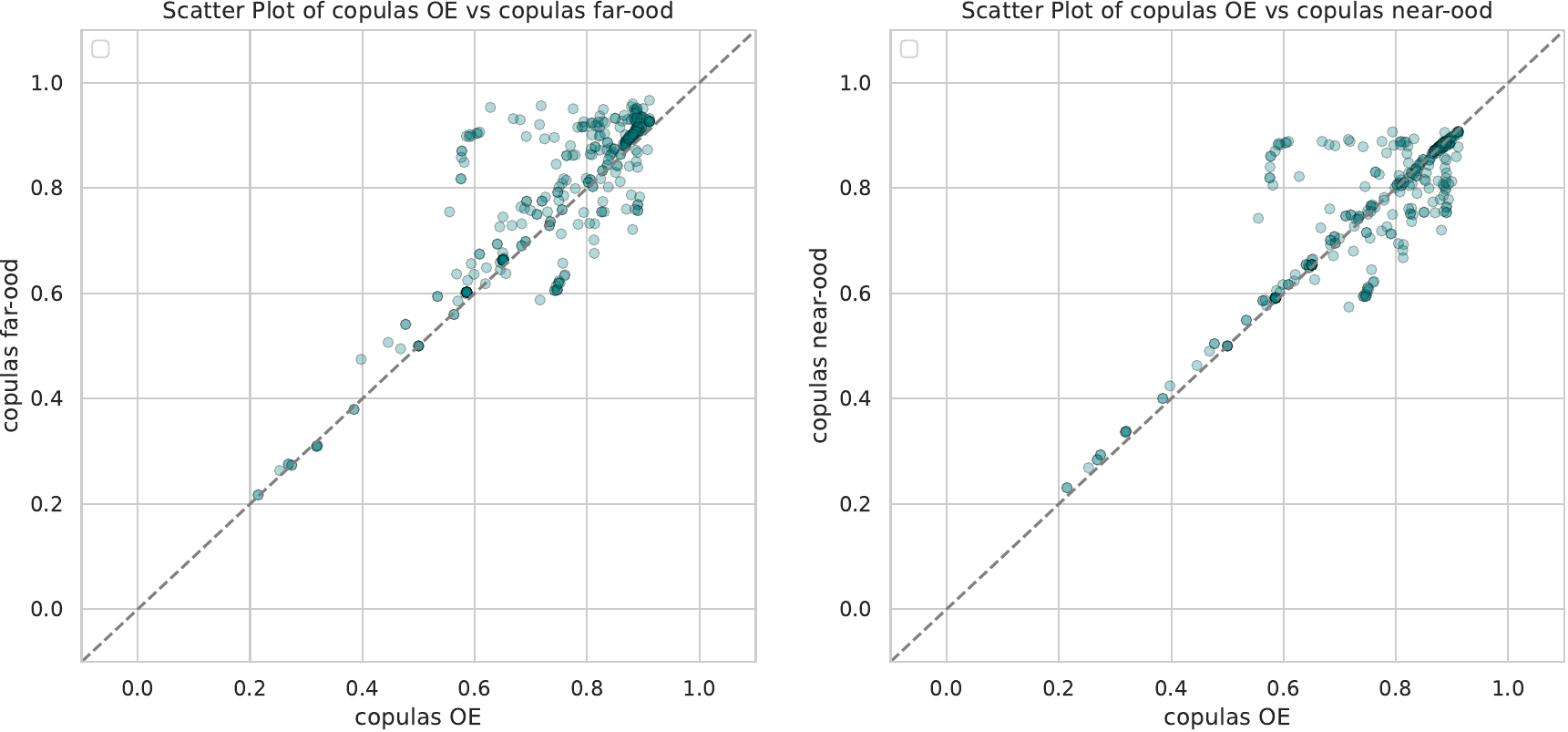}
    \caption{CIFAR-10: Copulas}
\end{figure*}

\begin{figure}[!h]
    \begin{subfigure}{0.49\textwidth}
    
          \label{fig:ccpv}
          \centering
          \includegraphics[width=1.0\textwidth, trim={2cm, 1cm, 2cm, 1cm}]{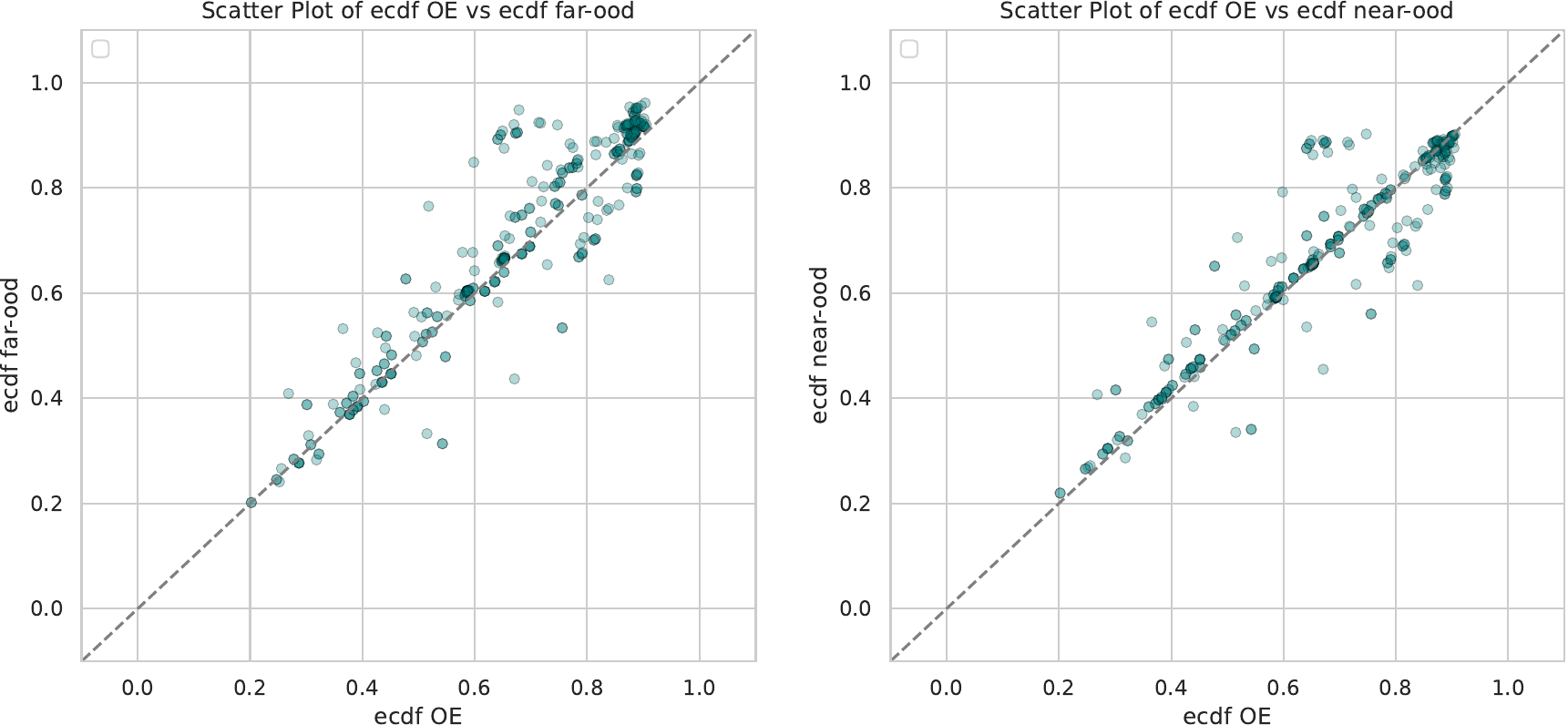}
            \caption{CIFAR-10: ECDF}
    \end{subfigure}
    \begin{subfigure}{0.49\textwidth}
          \label{fig:ccpv}
          \centering
          \includegraphics[width=1.0\textwidth, trim={2cm, 1cm, 2cm, 1cm}]{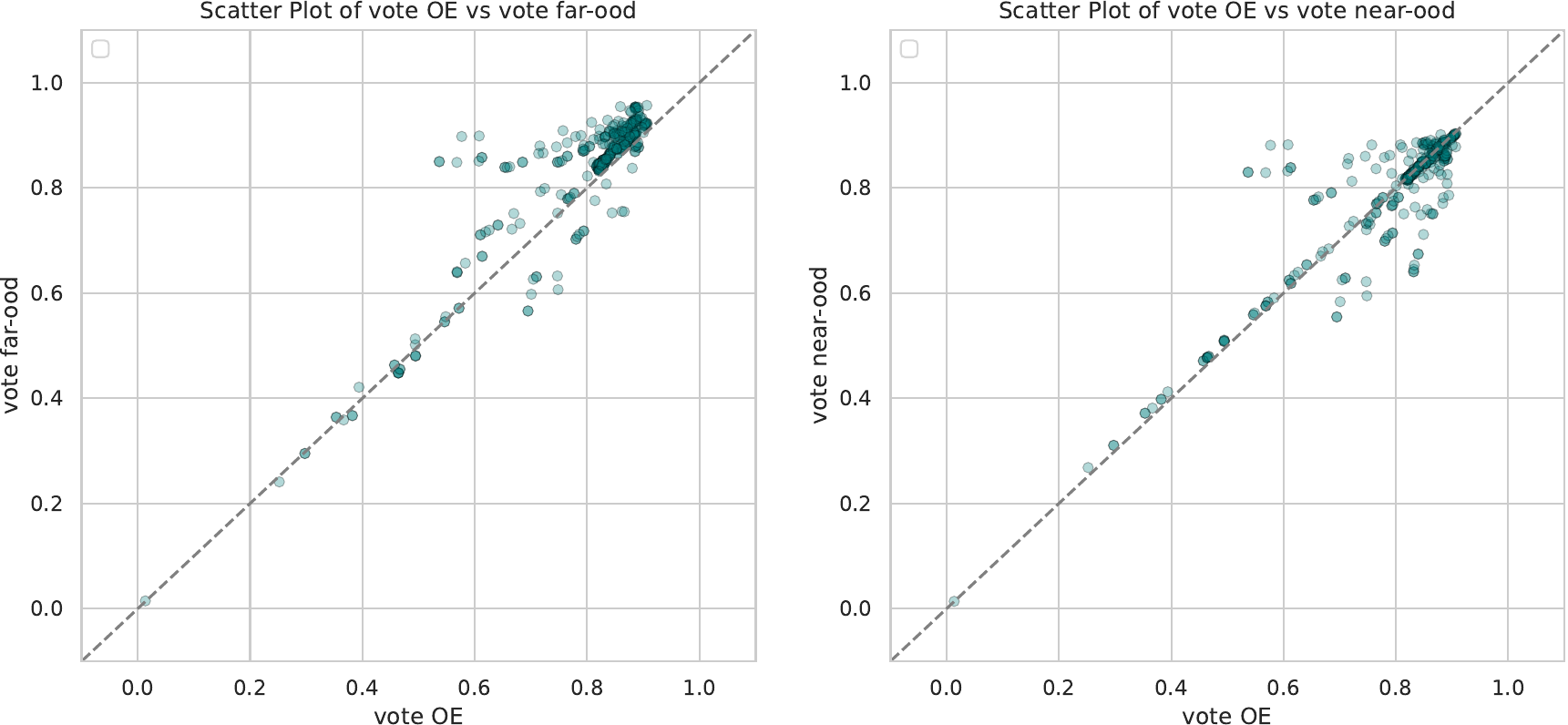}
            \caption{CIFAR-10: Majority Vote}
    \end{subfigure}
\end{figure}

\begin{figure}[!h]
    \begin{subfigure}{0.49\textwidth}
    
          \label{fig:ccpv}
          \centering
          \includegraphics[width=1.0\textwidth, trim={2cm, 1cm, 2cm, 1cm}]{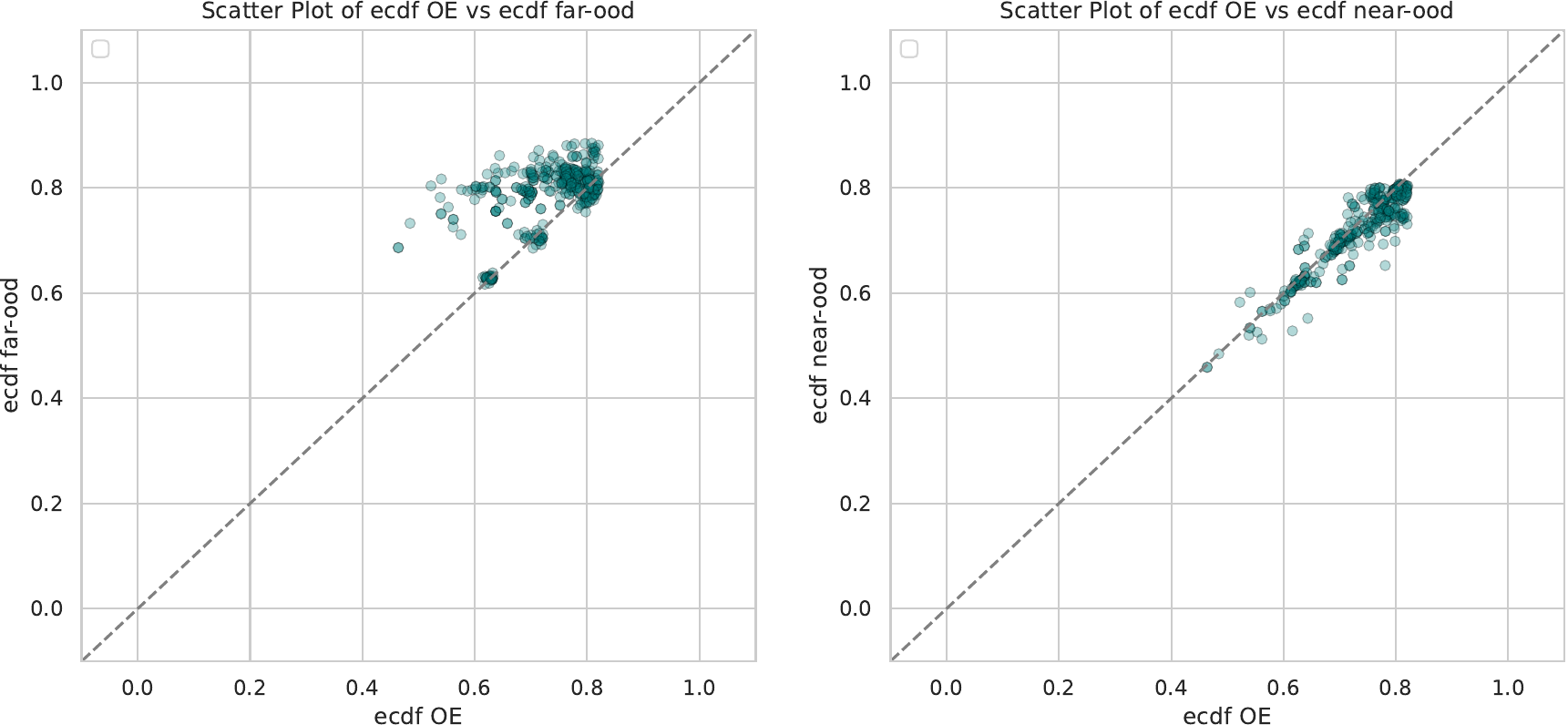}
            \caption{CIFAR-100: ECDF}
    \end{subfigure}
    \begin{subfigure}{0.49\textwidth}
          \label{fig:ccpv}
          \centering
          \includegraphics[width=1.0\textwidth, trim={2cm, 1cm, 2cm, 1cm}]{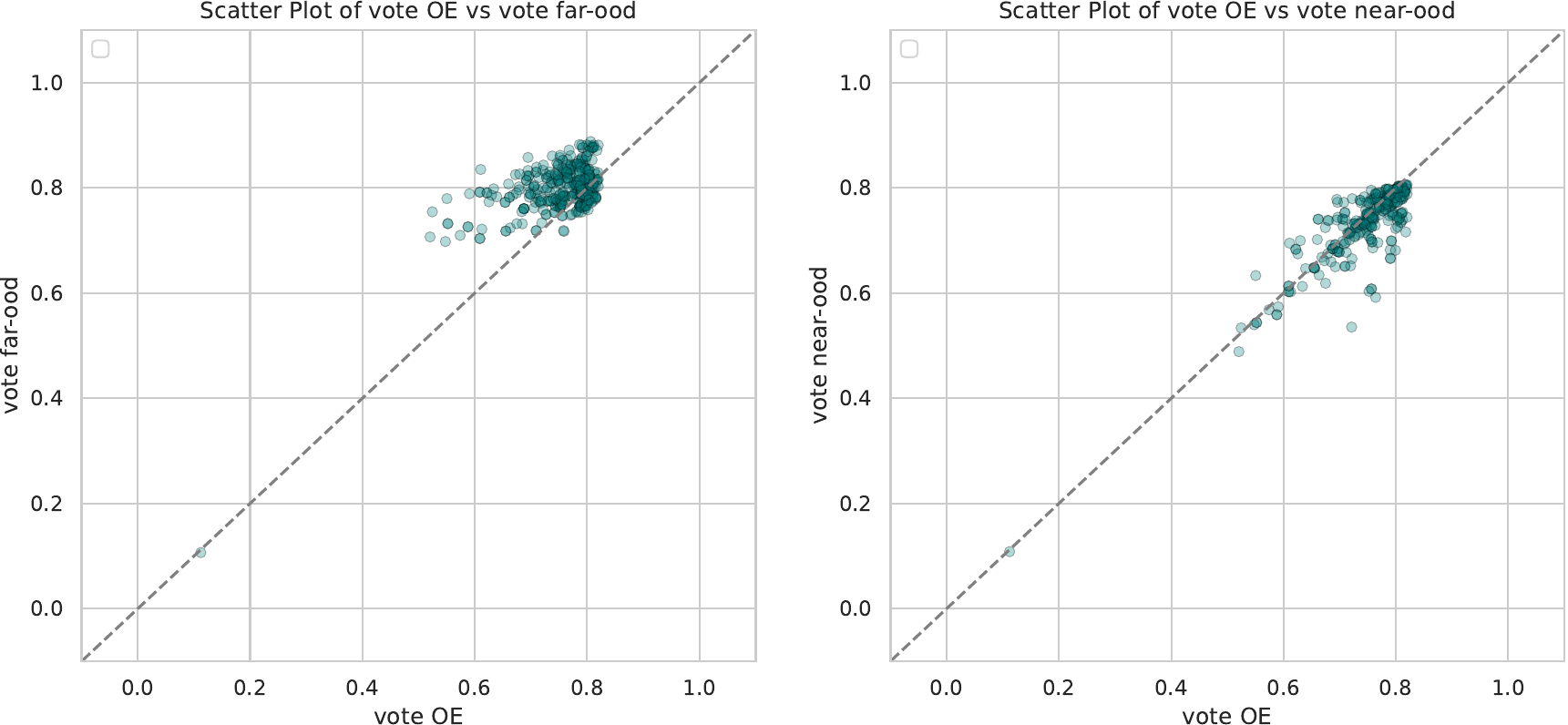}
            \caption{CIFAR-100: Majority Vote}
    \end{subfigure}
\end{figure}

\begin{figure}[!h]
    \begin{subfigure}{0.49\textwidth}
    
          \label{fig:ccpv}
          \centering
          \includegraphics[width=1.0\textwidth, trim={2cm, 1cm, 2cm, 1cm}]{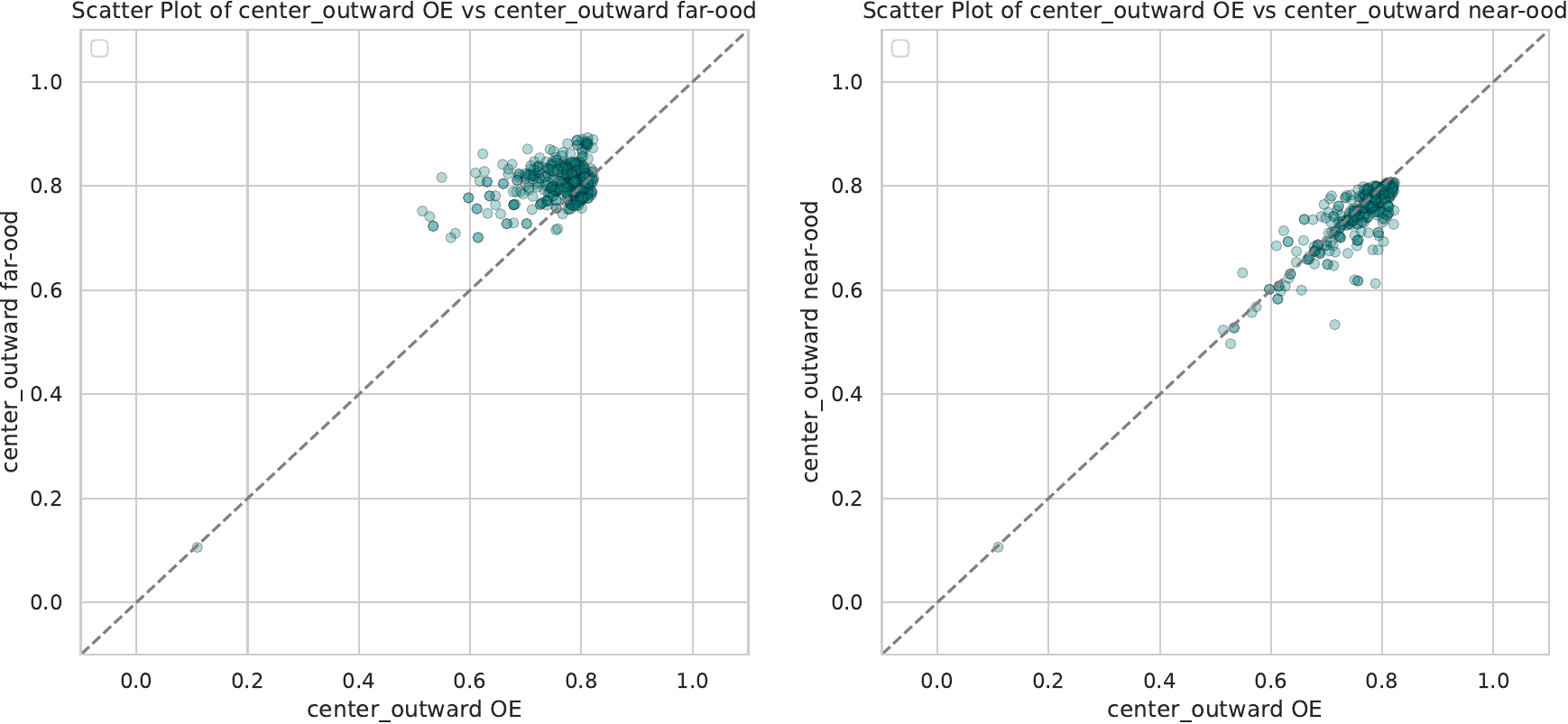}
            \caption{CIFAR-100: Center Outward}
    \end{subfigure}
    \begin{subfigure}{0.49\textwidth}
          \label{fig:ccpv}
          \centering
          \includegraphics[width=1.0\textwidth, trim={2cm, 1cm, 2cm, 1cm}]{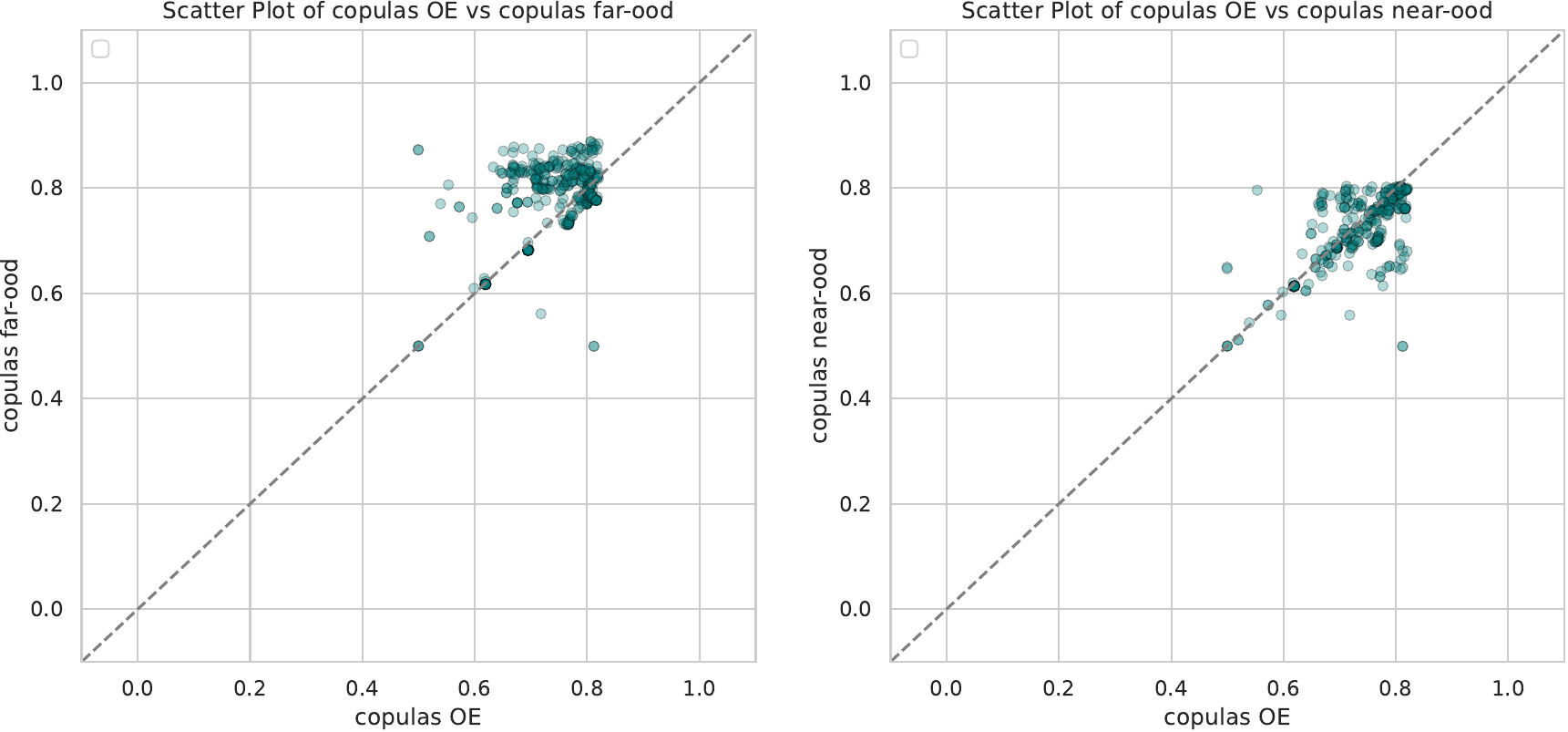}
            \caption{CIFAR-100: Copulas}
    \end{subfigure}
\end{figure}

\begin{figure}[!h]
    \begin{subfigure}{0.49\textwidth}
    
          \label{fig:ccpv}
          \centering
          \includegraphics[width=1.0\textwidth, trim={2cm, 1cm, 2cm, 1cm}]{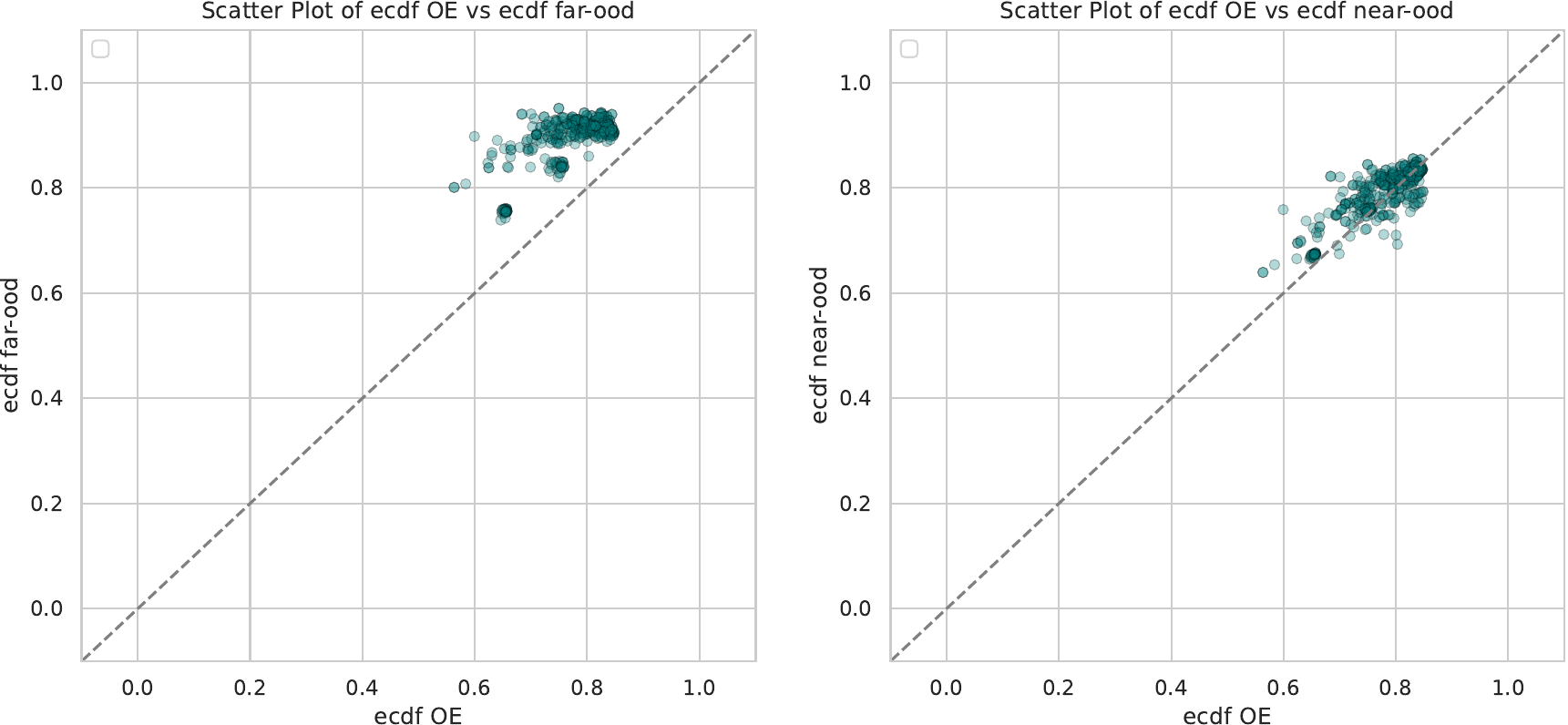}
            \caption{ImageNet-200: ECDF}
    \end{subfigure}
    \begin{subfigure}{0.49\textwidth}
          \label{fig:ccpv}
          \centering
          \includegraphics[width=1.0\textwidth, trim={2cm, 1cm, 2cm, 1cm}]{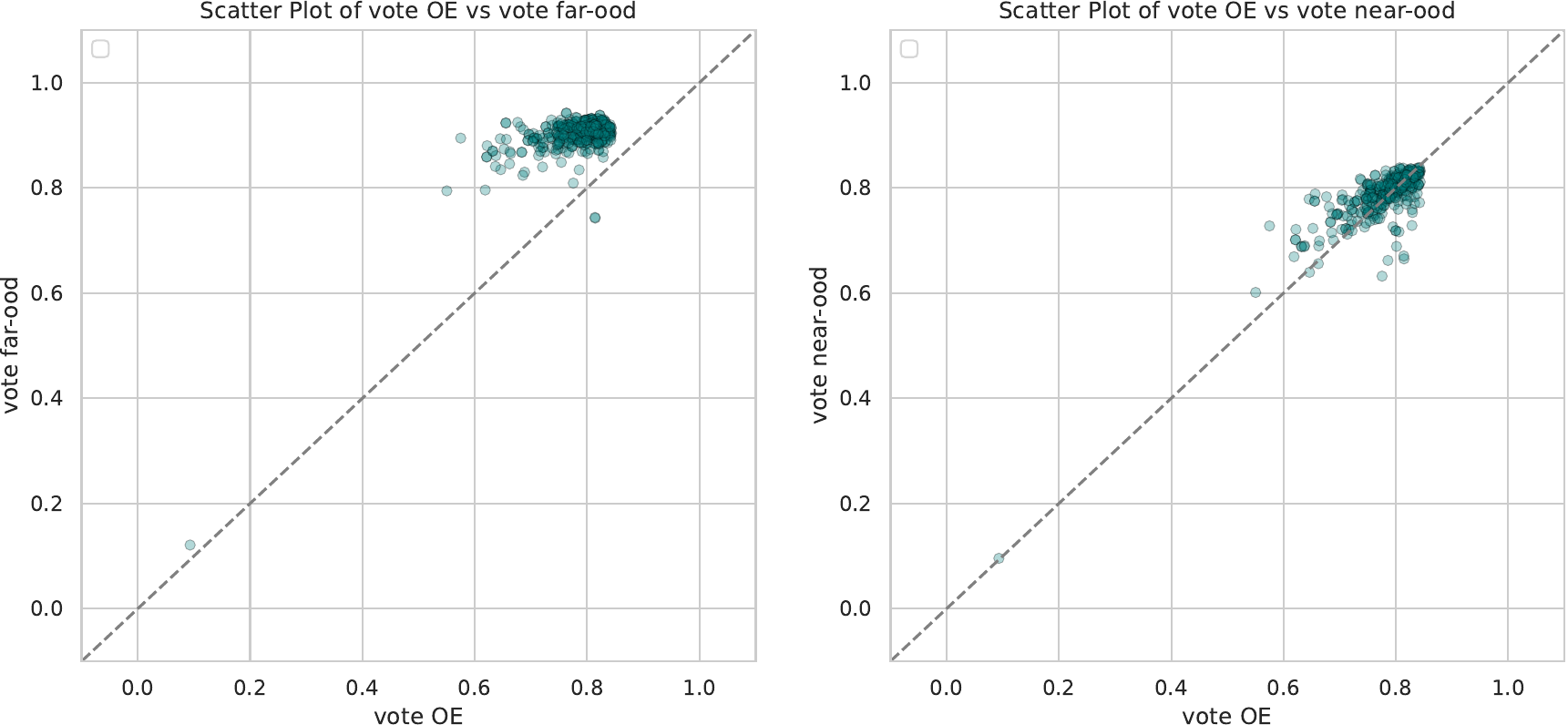}
            \caption{ImageNet-200: Majority Vote}
    \end{subfigure}
\end{figure}

\begin{figure}[!h]
    \begin{subfigure}{0.49\textwidth}
    
          \label{fig:ccpv}
          \centering
          \includegraphics[width=1.0\textwidth, trim={2cm, 1cm, 2cm, 1cm}]{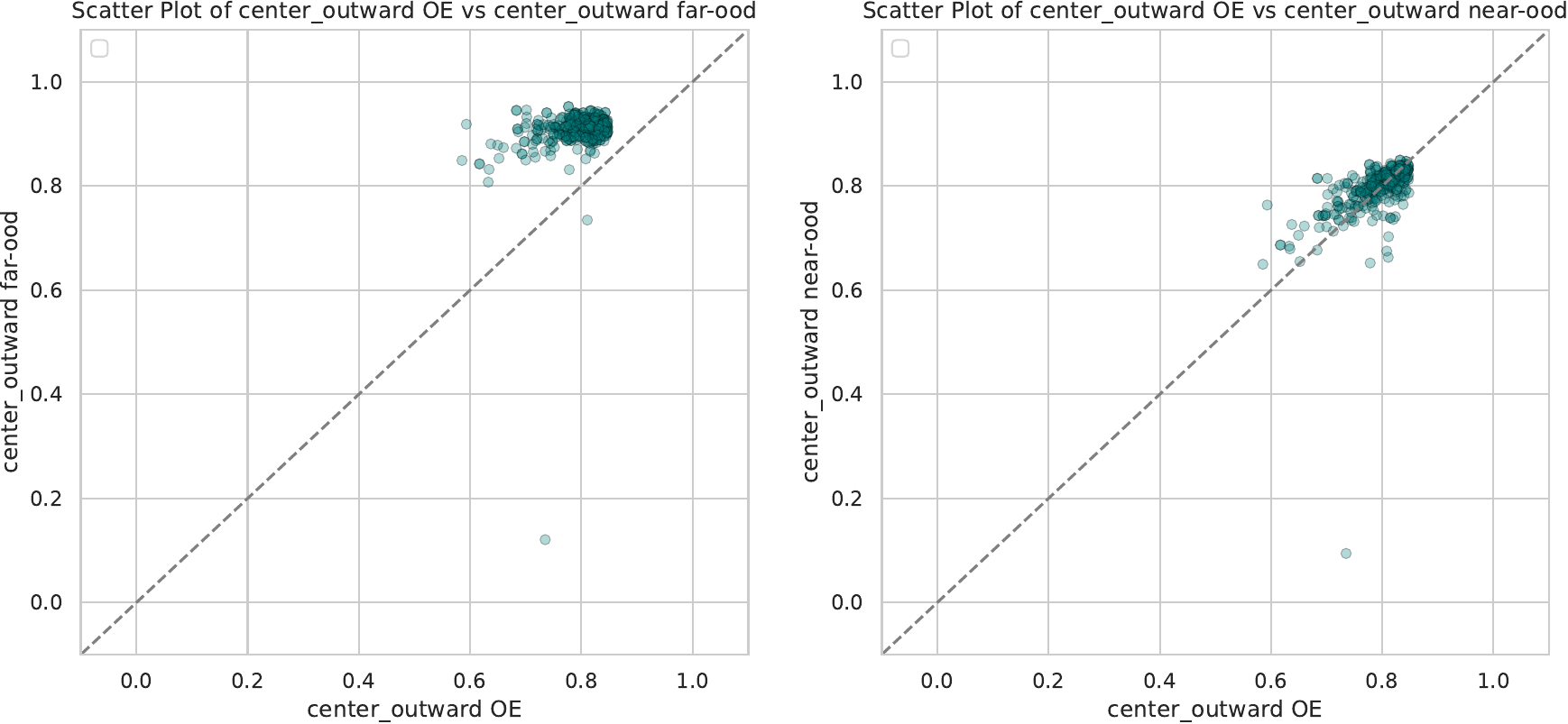}
            \caption{ImageNet-200: Center Outward}
    \end{subfigure}
    \begin{subfigure}{0.49\textwidth}
          \label{fig:ccpv}
          \centering
          \includegraphics[width=1.0\textwidth, trim={2cm, 1cm, 2cm, 1cm}]{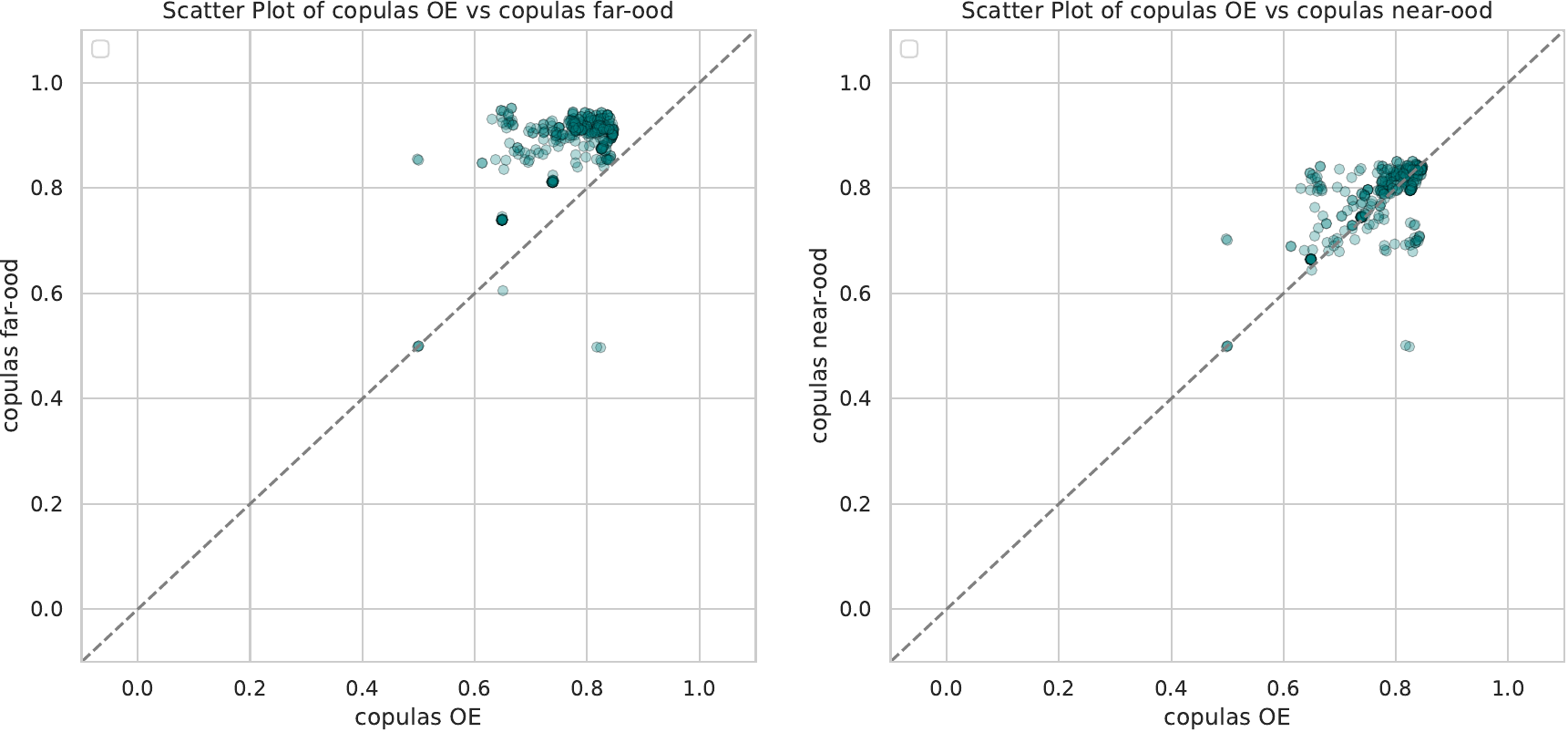}
            \caption{ImageNet-200: Copulas}
    \end{subfigure}
\end{figure}

\phantom{a}
\newpage
\section{Details on OpenOOD benchmark}\label{app:openood}

The OOD datasets used for the Near OOD and Far OOD clusters, and Outlier Exposure for each in-distribution dataset are:

\begin{itemize}
    \item CIFAR-10 as in distribution dataset: 
    \begin{itemize}
        \item Near OOD: CIFAR-100, Tiny ImageNet
        \item Far OOD: MNIST, SVHN, Textures, Places365.
        \item Outlier Exposure: Tiny597
    \end{itemize}
    \item CIFAR-100 as in distribution dataset: 
    \begin{itemize}
        \item Near OOD: CIFAR-10, Tiny ImageNet
        \item Far OOD: MNIST, SVHN, Textures, Places365.
        \item Outlier Exposure: Tiny597
    \end{itemize}
    \item Imagenet-200 as in distribution dataset: 
    \begin{itemize}
        \item Near OOD: SSB-hard, NINCO
        \item Far OOD: iNaturalist, Textures, Openimage-O
        \item Outlier Exposure: Imagenet-800
    \end{itemize}
\end{itemize}

For CFIAR-10 and CIFAR-100, a ResNet 18 is considered and for ImageNet-200, a ResNet 50.
 
\section{Complementary Results For Search Strategies With Outlier Exposure}\label{app:oe_results}

In this section, we present the results of the search strategies based on Outlier Exposure under the same form as Table \ref{tab:main_results}. To obtain these results, we select the candidate sets of OOD detectors returned by the procedure described in Section \ref{sec:oe} (highlighted in the Pareto plots), select the best on the validation dataset ($\setidval$, $\setoodval$), and then report its AUROC on OpenOOD's test dataset ($\setidtest$, $\setoodtest$). 

\begin{table*}[ht]
\centering
\resizebox{\textwidth}{!}{
\begin{tabular}{l|cc|cc|cc} 
    \toprule
     & \multicolumn{2}{c|}{\textbf{CIFAR-10}} & \multicolumn{2}{c|}{\textbf{CIFAR-100}} & \multicolumn{2}{c}{\textbf{ImageNet-200}} \\
    
    \midrule
         & {Near OOD} & {Far OOD}  & {Near OOD} & {Far OOD}  & {Near OOD} & {Far OOD}  \\
    \midrule
    Best indiv.               & 90.7 & {93.3} & {80.4}   & {86.6}  & {83.8}  & {93.0}  \\ 
    \midrule
    \multicolumn{7}{l}{\textbf{Majority Vote}} \\
    \midrule
    Top $5\%$ pairs OE        & {90.3}        & \textbf{95.6} & \textbf{80.6}   & \textbf{88.8}         & {83.8}         & \textbf{92.0} \\
    Sensitivity OE            & \textbf{91.3} & \textbf{95.6} & \textbf{82.2}   & \textbf{88.1}  & \textbf{86.5}  & {91.9} \\
    Beam Search OE            & 90.7 & \textbf{94.6} & \textbf{80.5} & 86.6 & \textbf{84.8} & \textbf{93.2} \\
    \midrule
    \multicolumn{7}{l}{\textbf{Empirical CDF}} \\
    \midrule
    Top $5\%$ pairs OE        & {90.2} & \textbf{96.1}        & \textbf{80.7}   & \textbf{88.4}  & \textbf{85.4}  & \textbf{94.0} \\
    Sensitivity OE            & {90.2} & {93.1}        & \textbf{82.1}   & {82.4}         & \textbf{86.9}  & {91.2} \\
    Beam Search OE            & 90.7 & 93.2 & 80.4 & \textbf{86.7} & 83.8 & 93.0\\
    \midrule
    \multicolumn{7}{l}{\textbf{Copulas} }\\
    \midrule
    Top $5\%$ pairs OE        & {90.7}        & \textbf{96.7} & \textbf{80.3}  & \textbf{88.9}  & \textbf{84.0}  & {92.2} \\
    Sensitivity OE            & \textbf{91.2} & \textbf{96.7} & \textbf{82.3}  & \textbf{88.4}  & \textbf{86.6}  & {92.6} \\
    Beam Search OE            & 90.7 & \textbf{93.4} & \textbf{80.5} & 86.6 & \textbf{84.5} & 93.0 \\
    \midrule
    \multicolumn{7}{l}{\textbf{Center Outward}} \\
    \midrule
    Top $5\%$ pairs OE        & {90.4} & \textbf{96.4} & \textbf{80.7}  & \textbf{89.2}  & \textbf{84.1}  & \textbf{92.6} \\
    Sensitivity OE            & {90.9} & \textbf{93.4} & \textbf{82.2}  & \textbf{88.9}  & \textbf{86.7}  & 92.6 \\
    Beam Search OE            & 90.7 & \textbf{95.9} & \textbf{80.7} & 86.6 & \textbf{85.1} & \textbf{93.2}\\
    \bottomrule
    \end{tabular}

}
\caption{AUROC of combination methods for best-performing sets of individual OOD detectors selected with search strategies. Top 5\% pairs OE - Selected among the top 5\% performing pairs evaluated on proxy OOD data (Outlier Exposure); Sensitivity OE - selected after a sensitivity analysis performed on proxy OOD data;  Beam Search - Selected after a Beam Search on proxy OOD data. }
\label{tab:openood_oe}
\end{table*}

We then report additional Pareto fronts with individual detectors and \multiooddetector{}s combined using different Combination methods, which are returned after different search strategies based on AUROCS obtained on Outlier Exposure datasets.

\begin{figure}[!h]
  \label{fig:pareto}
  \centering
  \includegraphics[width=1.0\textwidth]{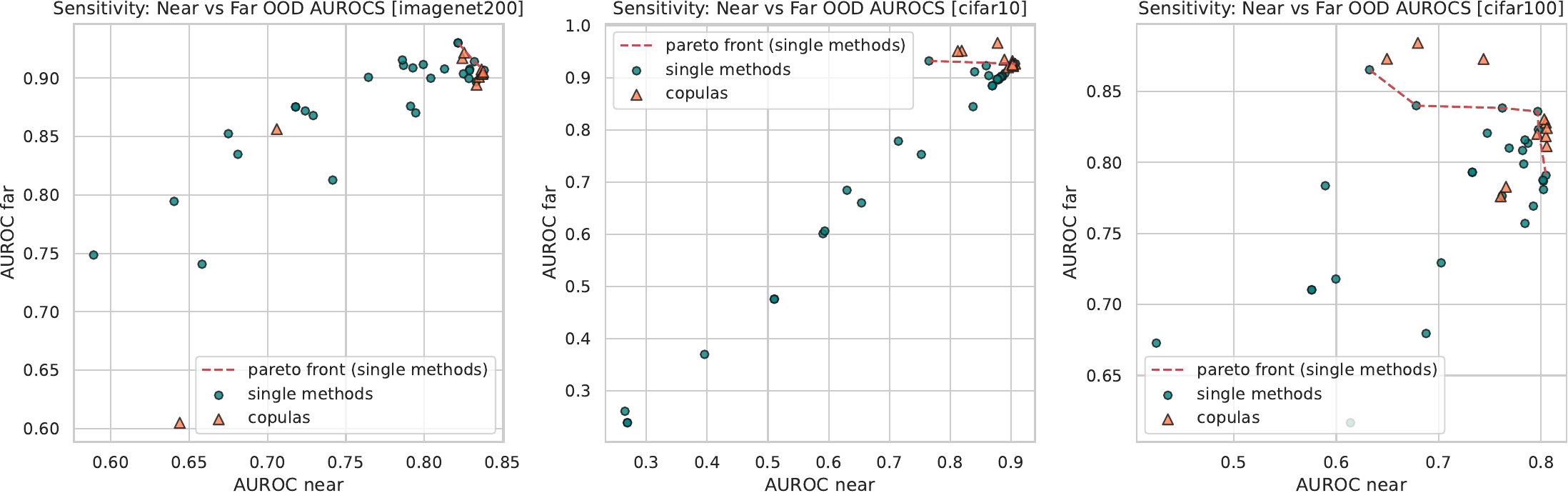}
    \caption{Copulas, Sensitivy}
\end{figure}

\begin{figure}[!h]
  \label{fig:pareto}
  \centering
  \includegraphics[width=1.0\textwidth]{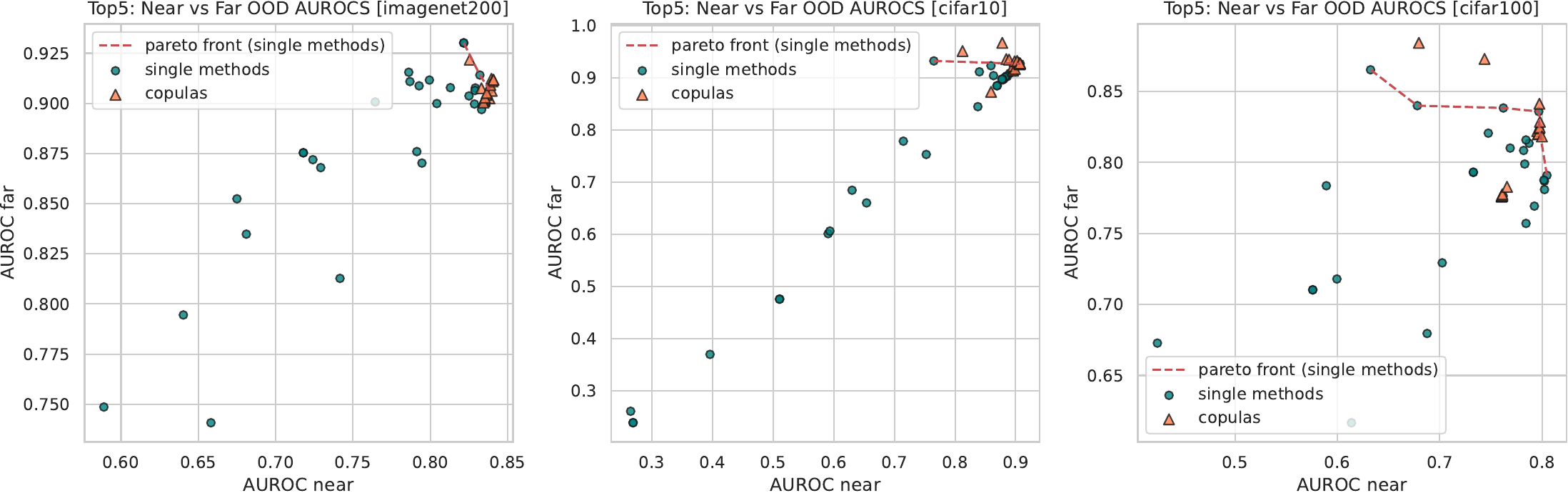}
    \caption{Copulas, Top $5\%$}
\end{figure}

\begin{figure}[!h]
  \label{fig:pareto}
  \centering
  \includegraphics[width=1.0\textwidth]{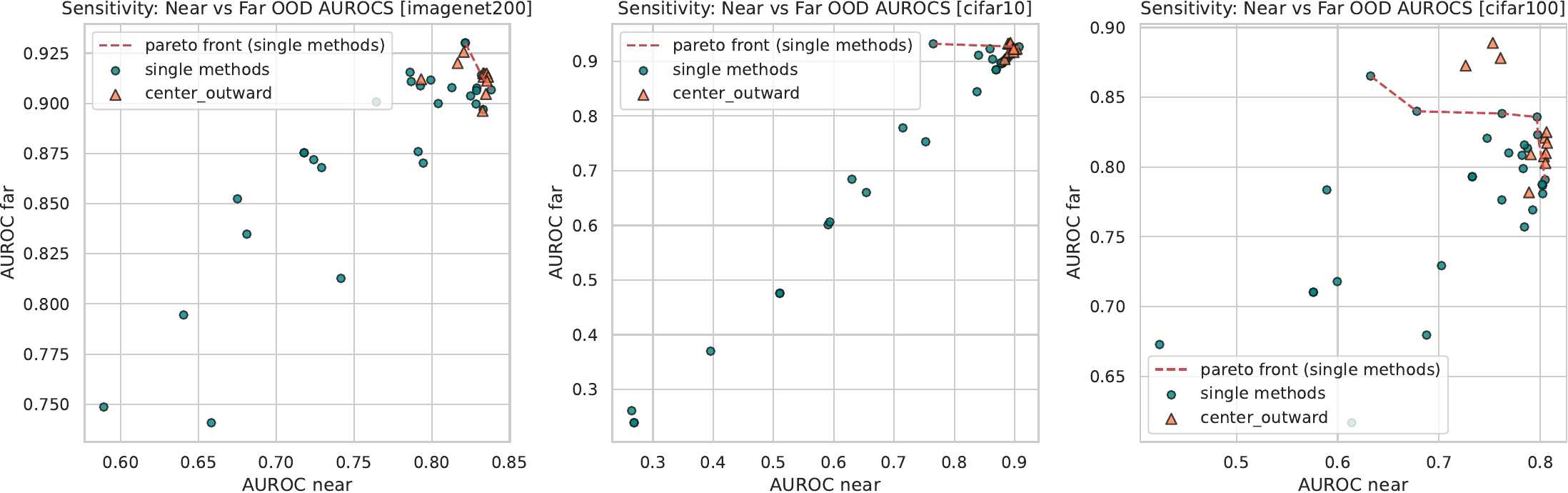}
    \caption{Center-Outward, Sensitivity}
\end{figure}

\begin{figure}[!h]
  \label{fig:pareto}
  \centering
  \includegraphics[width=1.0\textwidth]{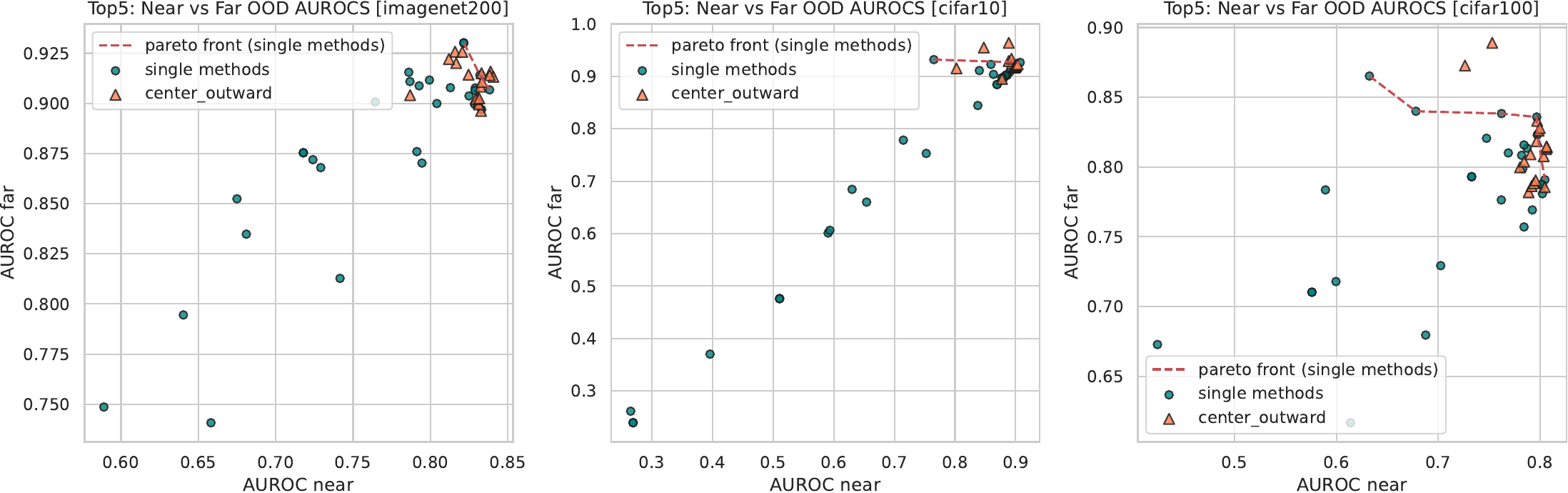}
    \caption{Center-Outward, Top $5\%$}
\end{figure}

\begin{figure}[!h]
  \label{fig:pareto}
  \centering
  \includegraphics[width=1.0\textwidth]{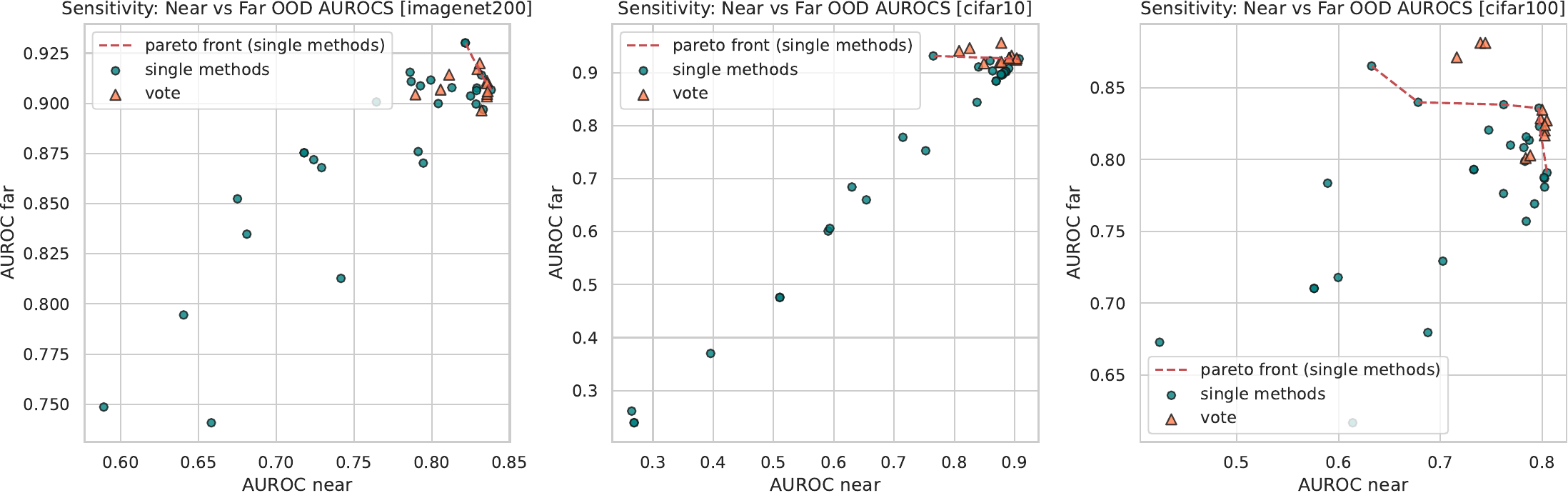}
    \caption{Majority Vote, Sensitivity}
\end{figure}

\begin{figure}[!h]
  \label{fig:pareto}
  \centering
  \includegraphics[width=1.0\textwidth]{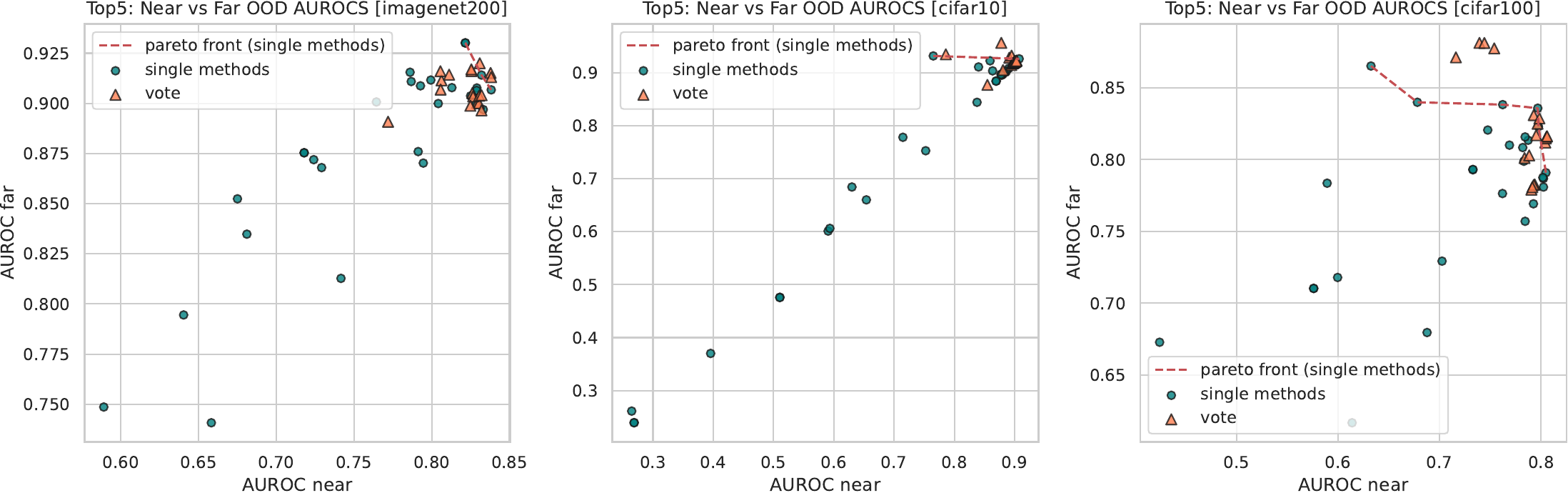}
    \caption{Majority Vote, Top $5\%$}
\end{figure}

\begin{figure}[!h]
  \label{fig:pareto}
  \centering
  \includegraphics[width=1.0\textwidth]{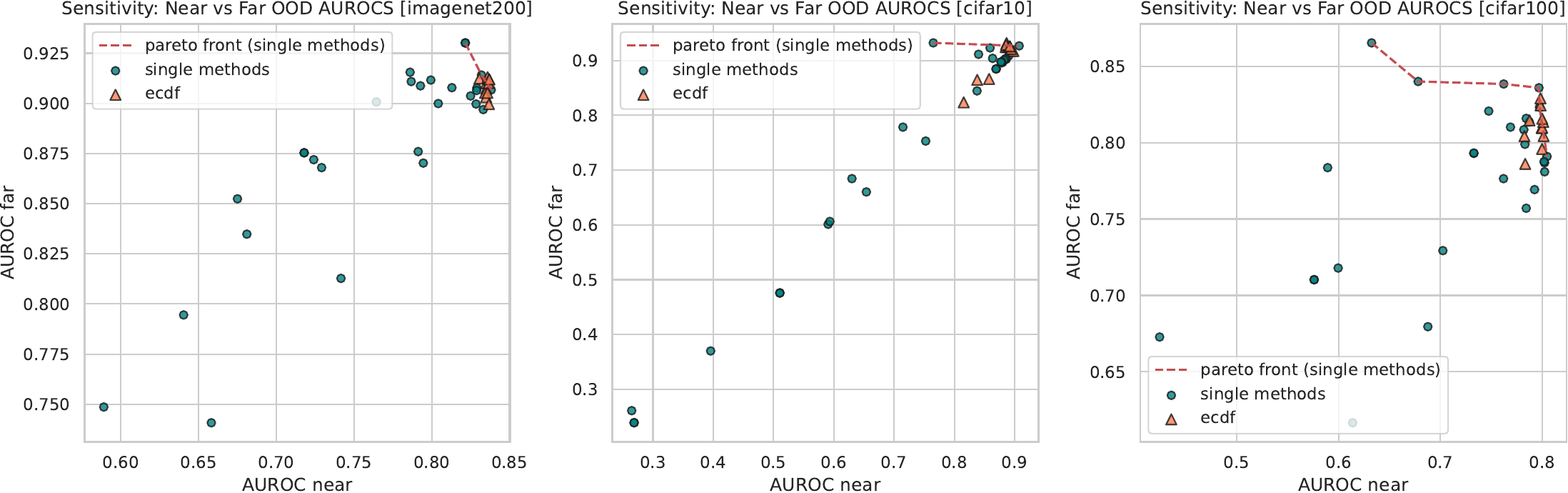}
    \caption{ECDF, Sensitivity}
\end{figure}

\begin{figure}[!h]
  \label{fig:pareto}
  \centering
  \includegraphics[width=1.0\textwidth]{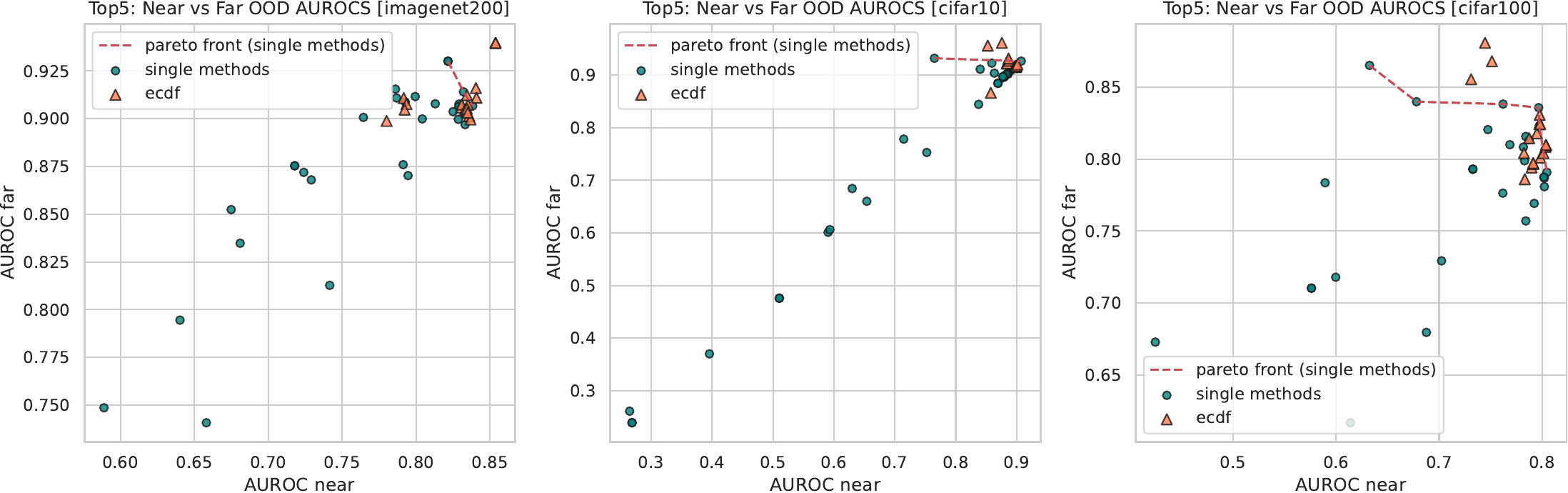}
    \caption{ECDF, Top $5\%$}
\end{figure}

\phantom{a}
\clearpage
\section*{NeurIPS Paper Checklist}
\begin{enumerate}

\item {\bf Claims}
    \item[] Question: Do the main claims made in the abstract and introduction accurately reflect the paper's contributions and scope?
    \item[] Answer: \answerYes{} 
    \item[] Justification: The contributions are faithfully summarized as a list in the introduction (see Section \ref{sec:intro}). The abstract further summarizes this list of contributions.

\item {\bf Limitations}
    \item[] Question: Does the paper discuss the limitations of the work performed by the authors?
    \item[] Answer: \answerYes{} 
    \item[] Justification: The paper contains a separate "Limitations" section (Section \ref{sec:limitations}) addressing the main limitations of our methods, namely data availability and computation resources.

\item {\bf Theory Assumptions and Proofs}
    \item[] Question: For each theoretical result, does the paper provide the full set of assumptions and a complete (and correct) proof?
    \item[] Answer: \answerNA{} 
    \item[] Justification: While we use mathematical objects in the proposed methods, we do not state any results nor theorems. 

    \item {\bf Experimental Result Reproducibility}
    \item[] Question: Does the paper fully disclose all the information needed to reproduce the main experimental results of the paper to the extent that it affects the main claims and/or conclusions of the paper (regardless of whether the code and data are provided or not)?
    \item[] Answer: \answerYes{} 
    \item[] Justification: Our experiments are carried out on well-known benchamrks using publicly available datasets, we only use publicly available pre-trained neural networks, which we provide references to in the text. 
    The code to replicate our post-processing methods has been made publicly available. 
    There can be minor fluctuations in the results due to hardware or software differences, but this fluctuations are negligible when it comes to affecting the main claims and conclusions of the paper.

\item {\bf Open access to data and code}
    \item[] Question: Does the paper provide open access to the data and code, with sufficient instructions to faithfully reproduce the main experimental results, as described in supplemental material?
    \item[] Answer: \answerYes{} 
    \item[] Justification: All data used in the paper is publicly available data, as well as all the pre-trained neural networks used in the experiments. On top of that, we provide the code for the post-processing techniques that we propose in the following (anonymized) github repository: \url{https://anonymous.4open.science/r/multi-ood-ECEC}

\item {\bf Experimental Setting/Details}
    \item[] Question: Does the paper specify all the training and test details (e.g., data splits, hyperparameters, how they were chosen, type of optimizer, etc.) necessary to understand the results?
    \item[] Answer: \answerYes{} 
    \item[] Justification: All the training and test details are provided. The most important details (for example the different benchmark datasets and data splits) are provided in the core of the paper (Section \ref{sec:experiments}). Additional details like hardware used, hyperparemeters, exact size of the data splits etc. are provided in Appendix \ref{app:comb} and Appendix \ref{app:openood}.

\item {\bf Experiment Statistical Significance}
    \item[] Question: Does the paper report error bars suitably and correctly defined or other appropriate information about the statistical significance of the experiments?
    \item[] Answer: \answerNo{} 
    \item[] Justification: Our experiments are carried out on benchmarks where the original scores do not carry any information about the statistical significance. There is very little randomness involved in our results (only that of the data splits, which have been obtained using a common random seed, 42).

\item {\bf Experiments Compute Resources}
    \item[] Question: For each experiment, does the paper provide sufficient information on the computer resources (type of compute workers, memory, time of execution) needed to reproduce the experiments?
    \item[] Answer: \answerYes{} 
    \item[] Justification: The details for the former are given in the introduction of the Apppendix~\ref{app:comb}.
    
\item {\bf Code Of Ethics}
    \item[] Question: Does the research conducted in the paper conform, in every respect, with the NeurIPS Code of Ethics \url{https://neurips.cc/public/EthicsGuidelines}?
    \item[] Answer: \answerYes{} 
    \item[] Justification: Our research does not involve human subjects nor participants, and all used data is publicly available data which is commonly used for OOD detection benchmarking purposes.

\item {\bf Broader Impacts}
    \item[] Question: Does the paper discuss both potential positive societal impacts and negative societal impacts of the work performed?
    \item[] Answer: \answerNo{} 
    \item[] Justification: Our work has the same positive and negative societal impact as any generic method for enhancing Out-of-Distribution Detection.
    
\item {\bf Safeguards}
    \item[] Question: Does the paper describe safeguards that have been put in place for responsible release of data or models that have a high risk for misuse (e.g., pretrained language models, image generators, or scraped datasets)?
    \item[] Answer: \answerNA{} 
    \item[] Justification: Our paper poses no such risks.

\item {\bf Licenses for existing assets}
    \item[] Question: Are the creators or original owners of assets (e.g., code, data, models), used in the paper, properly credited and are the license and terms of use explicitly mentioned and properly respected?
    \item[] Answer: \answerYes{} 
    \item[] Justification: We duely credit the sources for all dataset and pre-trained models we use as specified by the guidelines. 

\item {\bf New Assets}
    \item[] Question: Are new assets introduced in the paper well documented and is the documentation provided alongside the assets?
    \item[] Answer: \answerNA{} 
    \item[] Justification: The paper does not release new assets.

\item {\bf Crowdsourcing and Research with Human Subjects}
    \item[] Question: For crowdsourcing experiments and research with human subjects, does the paper include the full text of instructions given to participants and screenshots, if applicable, as well as details about compensation (if any)? 
    \item[] Answer: \answerNA{} 
    \item[] Justification: Our paper does not involve crowdsourcing nor research with human subjects.

\item {\bf Institutional Review Board (IRB) Approvals or Equivalent for Research with Human Subjects}
    \item[] Question: Does the paper describe potential risks incurred by study participants, whether such risks were disclosed to the subjects, and whether Institutional Review Board (IRB) approvals (or an equivalent approval/review based on the requirements of your country or institution) were obtained?
    \item[] Answer: \answerNA{} 
    \item[] Justification: Our paper does not involve crowdsourcing nor research with human subjects.

\end{enumerate}
\end{document}